\documentclass{article} %
\usepackage{iclr2027_conference,times}

\usepackage{amsmath,amsfonts,bm}

\def\eqref#1{equation~\ref{#1}}

\def\1{\bm{1}}

\DeclareMathAlphabet{\mathsfit}{\encodingdefault}{\sfdefault}{m}{sl}
\SetMathAlphabet{\mathsfit}{bold}{\encodingdefault}{\sfdefault}{bx}{n}

\def\tH{{\underline{\bm H}}}

\newtheorem{Remark}{Remark}

\usepackage{hyperref}
\usepackage{url}
\usepackage{booktabs}
\usepackage{algorithm}
\usepackage{pifont}
\usepackage{algorithmic}
\usepackage{graphicx,tcolorbox}
\usepackage{comment}
\usepackage{enumitem}
\usepackage{algorithm}
\usepackage{algorithmic}
\definecolor{orange}{RGB}{255,107,0}

\title{LLM Hallucination Detection via Low-rank Multilinear Decomposition}

\author{Devansh Sharma, Faizul Rakib Sayem, Shahana Ibrahim \\
Department of Electrical \& Computer Engineering, University of Central Florida \\
\texttt{\{devansh.sharma, faizulrakib.sayem, shahana.ibrahim\}@ucf.edu}}

\iclrfinalcopy
\begin{document}

\maketitle
\lhead{Preprint. Under review.}

\begin{abstract}

{Large language models (LLMs) frequently generate fluent yet factually incorrect responses, raising concerns about their reliability. Recent hallucination detectors increasingly exploit internal representations, but extracting informative signals from high-dimensional hidden states remains challenging. Existing approaches often rely on selected activations or pooling-based compression, potentially overlooking discriminative information distributed across Transformer layers, feature modes, and tokens. We propose \textbf{LOTUS} (\textbf{L}ow-rank \textbf{O}rder-statistic \textbf{TU}cker-based \textbf{S}ubspace learning), a multilinear low-rank decomposition framework for hallucination detection that extracts compact representations from LLM hidden states while reducing dependence on labeled data. Our approach first summarizes variable-length token representations using complementary statistical descriptors that capture peak activations, activation distributions, and cross-layer changes. We then leverage low-rank multilinear decomposition to extract compact representations that capture the underlying structure of LLM hidden states, which are subsequently used to train a lightweight hallucination classifier. Experiments across multiple LLMs and question-answering benchmarks demonstrate improved hallucination detection performance over recent state-of-the-art baselines.}

\end{abstract}

\section{Introduction}
\label{sec:intro}

{

Large language models (LLMs) have demonstrated remarkable capabilities in language understanding,
generation, and reasoning \citep{huang2023reasoning}, enabling widespread use in question answering,
summarization, and dialogue systems \citep{brown2020language,min2025multidim}. They nonetheless remain susceptible
to \emph{hallucination}, generating fluent and plausible responses that are factually incorrect or
unsupported by evidence \citep{huang2025survey,bang2025hallulens}. Detection is difficult because
hallucinated responses are often indistinguishable from correct ones at the surface level, remaining
coherent and linguistically well formed even when their content is incorrect
\citep{wang2024factuality,singh2025ubiquity}. Large reasoning models compound the problem by
producing logically structured reasoning traces that appear sound yet reach an incorrect conclusion
\citep{mei2026reasoning,yao2025reasoning}. 

Existing hallucination detection methods are broadly either output-based
\citep{farquhar2024detecting, manakul2023selfcheckgpt,zhang2023enhancing, shelmanov2025head} or
internal-representation-based
\citep{duan2024llms,hu2025harp,barshalom2025actvit,zeng2026halluguard}. The former often exploit uncertainty
or consistency across multiple generations, using signals such as lexical agreement
\citep{lin2022towards,manakul2023selfcheckgpt} or semantic entropy \citep{farquhar2024detecting}, and
therefore do not use the internal representations that give rise to those outputs. The latter treat
hidden states as a richer source of hallucination-related information, e.g., \citet{hu2025harp} projects
hidden representations onto subspaces derived from the model's unembedding matrix, and
\citet{barshalom2025actvit} models representations across layers
and tokens using a Vision Transformer (ViT)-inspired architecture. Both train supervised detectors on hallucination labels.

Despite their promise, hidden-state representations present several challenges for hallucination
detection. First, they are extremely high-dimensional, spanning multiple Transformer layers, response
tokens, and hidden features. Existing methods therefore reduce these representations before
classification, e.g., by selecting particular hidden states
\citep{hu2025harp,zeng2026halluguard} or by applying fixed-resolution pooling
\citep{barshalom2025actvit}. Such reductions make detection tractable but may discard complementary
information distributed across layers, tokens, and features. Directly learning from the full
hidden-state tensor avoids these reductions but results in a substantially higher-dimensional learning
problem, increasing computational cost and the risk of overfitting under limited supervision. Second,
response lengths vary, so the number of hidden states along the token mode differs across generations,
making a common fixed-size representation nontrivial. Third, hidden-state representations are
model-specific. LLMs differ in both the number of layers and hidden dimensions, making detectors
defined directly in the original hidden space architecture-dependent and difficult to transfer across
LLMs. %

{\bf Our contributions.}
We address these challenges with a framework based on a low-rank multilinear formulation of LLM hidden
representations. Our key contributions are:
\renewcommand{\labelitemi}{\ding{111}}
\begin{itemize}[leftmargin=*]
    \item \textit{Structured low-rank hidden-state representation.}
    We introduce \textbf{LOTUS} (\textbf{L}ow-rank \textbf{O}rder-statistic \textbf{TU}cker-based
    \textbf{S}ubspace learning) for LLM hallucination detection, which converts variable-length token
    representations into complementary layer-by-feature summaries and applies an unsupervised
    Tucker-2 projection \citep{tucker1966,kolda2009tensor} over the layer and feature modes,
    preserving the multilinear structure of the hidden states.
    \item \textit{Compact and efficient detection pipeline.}
    \textbf{LOTUS} learns its multilinear representation from unlabeled hidden states, with hallucination
    labels used only by a lightweight classifier. The low-rank decomposition reduces representation
    size and memory cost by mapping model-specific hidden states into fixed-dimensional cores, which
    also facilitates detector reuse across LLMs with different internal dimensions.
    \item \textit{Strong empirical performance with comprehensive evaluation.}
    We evaluate \textbf{LOTUS} across two LLMs and four question-answering benchmarks, where it
    delivers strong and consistent detection performance, and report ablations on the statistical
    summaries, Tucker ranks, projection design, performance under limited labeled
    data, and cross-model transfer.

\end{itemize}

\section{Problem Statement}
\label{sec:problem}

Let $\mathcal{M}$ be a decoder-only language model with $L$
transformer layers, each with hidden dimension $D$. Consider prompts $\{\bm x_i\}_{i=1}^{N}$, each associated with a reference answer $\bm a_i^*$, and for
each prompt sample $K$ responses ${\cal A}_i = \{\bm a_{i,1},\dots,\bm a_{i,K}\}$ from $\mathcal{M}$.
Let $s(\bm a_{i,k},\bm a_i^{*})$ denote the similarity between a generated answer and the reference
answer. Given a similarity threshold $\lambda$, the hallucination label of the $k$-th response is
\[
z_{i,k}
=
\mathbb{I}\!\left[s(\bm a_{i,k},\bm a_i^*)<\lambda\right],
\]
where $\mathbb{I}[\cdot]$ is the indicator function, so that $z_{i,k}=1$ denotes a hallucinated response
and $z_{i,k}=0$ a correct one. The goal of hallucination detection is to learn a scoring function
$c(\bm a_{i,k})$ that assigns larger values to hallucinated responses, from which a binary prediction
follows as
\[
g_{\tau}(\bm a_{i,k})
=
\mathbb{I}\!\left[c(\bm a_{i,k})\geq\tau\right],
\]
where $\tau$ is a decision threshold, with $g_{\tau}(\bm a_{i,k})=1$ denoting a hallucinated response
and $g_{\tau}(\bm a_{i,k})=0$ a non-hallucinated one.

}

{\bf Hidden-state tensor.}
Recent studies provide growing evidence that LLM internal representations encode cues relevant to hallucination detection across Transformer layers and token positions
\citep{azaria2023internal,chen2024inside,orgad2024llms,zhang2025detecting,barshalom2025actvit}. Since the location and strength of these cues vary across responses and datasets, detectors that rely
on a few selected activations can miss useful information, which motivates modeling hidden
representations jointly across layers, tokens, and feature dimensions.

Accordingly, we consider the internal representations generated by $\mathcal{M}$ for each response. For
the $k$-th response to prompt $\bm x_i$, let $\bm h_{i,k,t}^{(\ell)}\in\mathbb{R}^{D}$ denote the
hidden representation of its $t$-th response token after layer $\ell$, with $\ell\in\{1,\ldots,L\}$ and
$t\in\{1,\ldots,T_{i,k}\}$, where $T_{i,k}$ is the token length of that response. Since $\mathcal{M}$
is autoregressive, $\bm h_{i,k,t}^{(\ell)}$ is computed conditioned on the prompt $\bm x_i$ and the
preceding generated tokens $\bm a_{i,k,<t}$. Collecting these representations across layers and answer tokens yields the
\textit{hidden-state tensor} given by
\[
\tH_{i,k}
\in
\mathbb{R}^{L\times T_{i,k}\times D},
\]
with $\tH_{i,k}[\ell,t,:] = \bm h_{i,k,t}^{(\ell)}$, whose entries preserve three complementary modes of
variation: Transformer layer, token position, and hidden feature dimension. Using these representations for hallucination detection poses three practical challenges:
\textit{(i)} the hidden-state tensor is extremely high-dimensional, so direct processing is memory-
and computation-intensive and requires substantially more labeled data to avoid overfitting;
\textit{(ii)} the token dimension $T_{i,k}$ varies across responses, requiring a fixed-size
representation without token alignment; and \textit{(iii)} $L$ and $D$ vary across LLMs, so detectors
defined directly in the original hidden space do not transfer readily across models.
This raises a central research question:

\begin{tcolorbox}[colback=gray!15, colframe=gray!15]
\centering
\textit{How can we extract compact, informative hidden-state representations that enable computationally efficient and label-efficient hallucination detection while facilitating transfer across LLMs?}
\end{tcolorbox}

\vspace{-0.3cm}
\section{Proposed approach}
\label{sec:method}

{ Our approach is motivated by the hypothesis that hallucination-discriminative information can be captured
by compact, low-rank multilinear representations of LLM hidden states. Rather than learning a
detector directly from the high-dimensional hidden-state tensor, we propose a two-stage framework that
separates representation learning from hallucination classification. The first stage constructs
fixed-size statistical summaries of hidden states and learns label-free Tucker-2 projections over the layer
and feature modes to obtain compact response representations. The second stage concatenates the
resulting low-dimensional representations and trains a lightweight hallucination classifier.
Fig.~\ref{fig:method} gives the complete overview of \textbf{LOTUS}, and we describe each component
below. 

\begin{figure}[t]
\centering
\includegraphics[width=\linewidth]{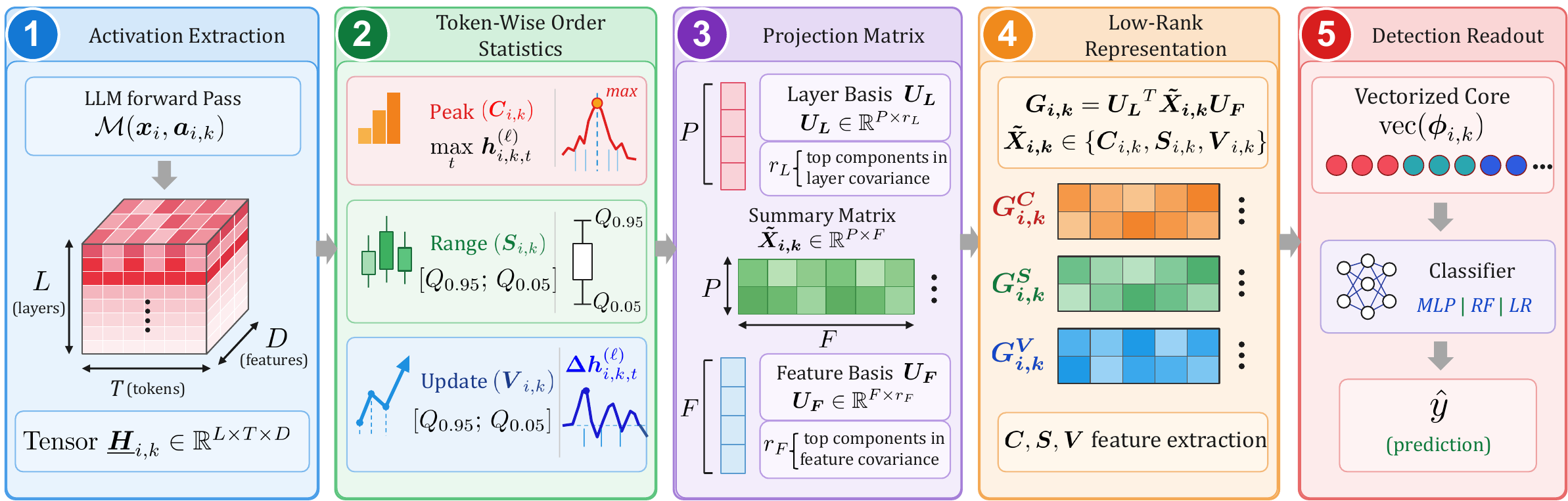}
\caption{Overview of our proposed framework \textbf{LOTUS}. Hidden states are summarized across tokens using peak activations, activation quantiles, and cross-layer differences, projected with summary-specific Tucker-2 bases learned from unlabeled activations, and concatenated for lightweight hallucination classification.}
\label{fig:method}
\vspace{-0.4cm}
\end{figure}

{\bf Order statistics over the token mode.}
\label{sec:summaries}
The hidden-state tensor $\tH_{i,k}\in\mathbb{R}^{L\times T_{i,k}\times D}$ varies in size along
the token mode, and hallucination-related activations may appear at different token positions. Hence,
representations taken from predefined positions may overlook informative signals. We
therefore summarize the token dimension using complementary statistical descriptors that produce
fixed-size representations while retaining information about activation magnitude, distribution, and
cross-layer changes. For each response $\bm a_{i,k}$ and each selected layer $\ell$, we construct three descriptor matrices:
\begin{align*}
    \bm C_{i,k}[\ell,:] &= \big(\max_t~\{\bm h^{(\ell)}_{i,k,t}\}\big)^{\top}, ~~~\text{({\bf Peak activation matrix})}\\
\bm S_{i,k}[\ell,:] &= \big(\mathrm{Q}_{0.95}(\{\bm h^{(\ell)}_{i,k,t}\}_t)^{\top},\;
\mathrm{Q}_{0.05}(\{\bm h^{(\ell)}_{i,k,t}\}_t)^{\top}\big), ~~~\text{({\bf Activation quantile matrix})}\label{eq:range}\\
\bm V_{i,k}[\ell,:] &= \big(\mathrm{Q}_{0.95}(\{\Delta\bm h^{(\ell)}_{i,k,t}\}_t)^{\top},\;
\mathrm{Q}_{0.05}(\{\Delta\bm h^{(\ell)}_{i,k,t}\}_t)^{\top}\big),~~~\text{({\bf Cross-layer difference matrix})}
\end{align*}
where $\max_t~\{\bm h_{t}\}\in\mathbb{R}^D$ and $Q_{\alpha}(\{\bm h_t\}_t)\in\mathbb{R}^D$ denote the
element-wise maximum and the element-wise empirical $\alpha$-quantile of the hidden-state vectors over
the token dimension, and the cross-layer difference
$\Delta\bm h^{(\ell)}_{i,k,t}=\bm h^{(\ell+1)}_{i,k,t}-\bm h^{(\ell)}_{i,k,t}$ is the change in the
hidden states between consecutive layers. 

The three summaries capture complementary properties of hidden representations. The peak activation
matrix captures localized activation peaks irrespective of their positions, since hallucination-related
patterns may concentrate in a few tokens whose positions vary across responses
\citep{orgad2024llms,barshalom2025actvit}. Because a maximum depends on a single token, the activation
quantile matrix characterizes how feature activations are distributed across the response,
distinguishing isolated peaks from sustained patterns. The cross-layer difference matrix summarizes how
representations evolve as successive Transformer layers process them \citep{zhang2025icr}. Together,
these position-independent summaries turn variable-length token representations into fixed-size
matrices without token alignment.
}

{

{\bf Unsupervised Tucker-2 projection over layer and feature modes.}
\label{sec:tucker}
The token summaries are of fixed-size but remain high-dimensional, particularly along
the feature mode. To extract compact representations that capture dependencies across
layers and hidden features, we employ a \textit{Tucker-2 decomposition}
\citep{tucker1966,kolda2009tensor}, which learns a compact, separable projection comprising a layer
basis and a feature basis, shared across all training responses and learned without any hallucination
labels. Projecting each summary matrix onto these subspaces yields a low-dimensional core matrix that
retains interactions between the layer and feature dimensions.

Let $\bm X_{i,k} \in \mathbb{R}^{P \times F}$ denote any one of the summary matrices
$\bm C_{i,k}$, $\bm S_{i,k}$, or $\bm V_{i,k}$, where
$P$ is the layer-mode dimension and $F\in\{D,2D\}$ the feature-mode dimension. The layer dimension $P$
need not equal the total number of layers $L$---it can be set smaller to reduce computational cost, and
the choice of which layers to retain can incorporate prior knowledge (see more details about the layer selection in Section~\ref{sec:setup}). Before
decomposition, we robustly scale each summary ${\bm X}_{i,k}$ with scaling parameters estimated from
the training responses only, which prevents a small number of large-magnitude activations from
dominating the projection, and denote the scaled summary by $\tilde{\bm X}_{i,k}$
(more details about this scaling are provided in Appendix~\ref{app:setup}).
From these scaled hidden representations $\tilde{\bm X}_{i,k}$, we seek \textit{shared orthonormal bases}
$\bm U_L\in\mathbb{R}^{P\times r_L}$ and
$\bm U_F\in\mathbb{R}^{F\times r_F}$, with $r_L<P$ and $r_F<F$,
that maximize the energy retained after projection onto the
corresponding low-dimensional subspaces:
\begin{equation}
\begin{aligned}
\max_{\bm U_L,\bm U_F}\quad&
\sum_{i=1}^N
\sum_{k=1}^{K}
\left\|
\bm U_L^{\top}\tilde{\bm X}_{i,k}\bm U_F
\right\|_F^2\\
\text{s.t.}\quad&
\bm U_L^{\top}\bm U_L=\bm I,\qquad
\bm U_F^{\top}\bm U_F=\bm I.
\end{aligned}
\label{eq:tucker-objective}
\end{equation}
where $\bm I$ is an identity matrix of appropriate size. This is the Tucker-2 decomposition of the
third-order tensor formed by stacking the training summaries along a response mode
\citep{tucker1966,kolda2009tensor}. It compresses only the layer and feature modes,
preserving a separate low-dimensional representation for each response. 

We handle the optimization problem in \eqref{eq:tucker-objective} with truncated higher-order singular value decomposition
(HOSVD)
\citep{delathauwer2000hosvd}. Specifically, the layer basis
$\bm U_L$ consists of the $r_L$ leading eigenvectors of $\sum_{i=1}^N
\sum_{k=1}^K
\tilde{\bm X}_{i,k}\tilde{\bm X}_{i,k}^{\top}$,
while the feature basis $\bm U_F$ is obtained from the
$r_F$ leading eigenvectors of $\sum_{i=1}^N
\sum_{k=1}^{K}
\tilde{\bm X}_{i,k}^{\top}\tilde{\bm X}_{i,k}$,
We compute the feature eigenspace with randomized SVD on the vertically stacked rows of all training
summaries \citep{halko2011finding} rather than forming the $F\times F$ Gram matrix, which reaches
$F=8192$ for LLaMA-3.1-8B. 
Once the bases are estimated, each response is mapped to
a low-dimensional core matrix defined as
\begin{equation}
\bm G_{i,k}
=
\bm U_L^{\top}
\tilde{\bm X}_{i,k}
\bm U_F
\in\mathbb{R}^{r_L\times r_F},
\label{eq:core}
\end{equation}
where $\bm G_{i,k}$ contains $r_Lr_F$ coefficients that represent
interactions between the layer and feature directions. In this way, we learn a separate core representation for each summary type $\tilde{\bm X}_{i,k}$.

\begin{Remark}[Projection Parameter Efficiency]
For the same $r_Lr_F$-dimensional output, Tucker-2 represents the projection using separate
layer and feature bases, requiring only $Pr_L+Fr_F$ projection coefficients. Applying principal component
analysis (PCA) to the vectorized summaries instead learns a single projection on the joint
layer--feature space and requires $PFr_Lr_F$ coefficients. Across the three summary types, this corresponds to
approximately $1.15$M versus $45.6$M coefficients for Qwen2.5-7B-Instruct and $1.31$M versus
$52.2$M for LLaMA-3.1-8B. The smaller parameterization also lowers the cost of storing the projection.
\end{Remark}
}
{
{\bf Hallucination detector from low-dimensional representations.}
\label{sec:readout}
To learn the hallucination detector, we vectorize and concatenate the low-dimensional core
representations $\bm G^C_{i,k}$, $\bm G^S_{i,k}$, and $\bm G^V_{i,k}$ of $\bm a_{i,k}$ into the
\textit{response feature vector}:
\begin{equation}
\bm \phi_{i,k}
=
\begin{bmatrix}
\mathrm{vec}(\bm G^C_{i,k}),
\mathrm{vec}(\bm G^S_{i,k}),
\mathrm{vec}(\bm G^V_{i,k})
\end{bmatrix}^{\top}
\in\mathbb R^{d_\phi},
\label{eq:features}
\end{equation}
where
$d_\phi =
r_L^C r_F^C +
r_L^S r_F^S +
r_L^V r_F^V$
is the dimension of the resulting feature vector. We then train a lightweight classifier
$f_{\bm\theta}$, with parameters $\bm\theta$, on the feature vectors of the training responses and
their hallucination labels $z_{i,k}$. At inference, the statistical summaries of an unseen response
$\bm a_{\mathrm{test}}$ are computed from its hidden-state tensor, transformed with the learned projection bases
using \eqref{eq:core}, and assembled as in \eqref{eq:features} into $\bm\phi_{\mathrm{test}}$. The classifier
returns the hallucination score $c(\bm a_{\mathrm{test}}) = f_{\bm\theta}(\bm \phi_{\mathrm{test}})$, which can be thresholded at
$\tau$ to obtain the binary decision $g_{\tau}$. 

}

{ 

\section{Experiments}
\subsection{Evaluation Setup}\label{sec:setup}
\vspace{-0.3cm}

{\bf Models and datasets.} We evaluate on two open-weight models of comparable scale:
Qwen2.5-7B-Instruct, which is instruction-tuned, and the base LLaMA-3.1-8B. We use four question answering datasets: TruthfulQA \citep{lin2022truthfulqa}, TriviaQA \citep{joshi2017triviaqa}
without evidence documents, NQ-Open \citep{kwiatkowski2019natural}, and the English questions of
TyDiQA-GP \citep{clark2020tydi}, in which each question comes with a passage containing its answer. For each question we generate $K=10$ responses by beam sampling with $10$ beams, returning $10$
sequences each capped at $64$ new tokens. The datasets contain $817$, $9{,}960$, $3{,}610$, and $440$
questions, respectively. A question is \emph{model-known} if at least one of its $K$ responses is
labeled correct, and \emph{model-unknown} otherwise. Only model-known questions are eligible for
detector training; their number varies substantially across models and datasets, and
Qwen2.5-7B-Instruct has notably fewer on NQ-Open, which contributes to the larger variance in that
setting. Appendix~\ref{app:setup} provides the exact counts, prompts, generation parameters, and
reproducibility details. 

{\bf Hallucination labels.} We label responses following the procedure in \citep{hu2025harp}. A
response $\bm a_{i,k}$ is compared with the reference answers of question $i$ and labeled correct
($z_{i,k}=0$) if its ROUGE-L score \citep{lin2004rouge}, averaged over the references, is at least
$0.7$, or if its BLEURT-20 score \citep{sellam2020bleurt} exceeds $0.5$ for at least one reference;
otherwise it is labeled hallucinated ($z_{i,k}=1$).  All methods are evaluated on the same
generated responses and labels. Appendix~\ref{app:labels} compares these labels with those of the judge
released by \citet{hu2025harp}.

{\bf Train--test splitting protocol.}
\label{sec:splits}
Since each
question is associated with multiple generated responses, the set of question-response pairs can be divided by question or by response. We use
the \emph{question-level} train-test split throughout, where we randomly assign $75\%$ of the model-known questions
    to training and the remaining $25\%$ to testing,
    keeping all responses to each question in the same
    split. The test set additionally includes all responses
    to model-unknown questions. Consequently, no question
    appears in both the training and test sets. The \emph{answer-level}
alternative pools responses instead, which puts responses to the same question on both train and test splits and
evaluates a detector on questions it was trained on (e.g., HARP \citet{hu2025harp} follows this protocol in their released implementation). {Appendix~\ref{app:split}} reports both splits and demonstrates how question overlap can inflate reported hallucination detection performance. The choice affects only detectors that score individual responses, since one producing a
single score per question is unaffected.

{ 

{\bf Evaluation metrics.}
\label{sec:reporting}
We use pooled AUROC as our primary metric, following standard practice in hallucination detection
\citep{hu2025harp,du2024haloscope,barshalom2025actvit}. Pooled AUROC compares responses across
different questions, so it can reward a detector for identifying questions that are generally more
prone to hallucination rather than for separating correct and hallucinated responses to the same
question. We therefore also report \emph{within-question AUROC}, which restricts the pairwise
comparison to responses of the same question and so measures response-level discrimination while
controlling for question-specific effects. A detector that assigns an identical score to every
response of a question obtains $50\%$, however high its pooled AUROC. It is closely related to the within-question discrimination of
\citet{taubenfeld-etal-2025-confidence}. Question- and dataset-level shortcuts in hallucination
detection have likewise been reported \citep{dubanowska2025representation,seo2025question}.
Within-question AUROC is defined only for questions containing at least one correct and one
hallucinated response---more details in Appendix~\ref{app:within}.

}

{\bf LOTUS implementation details.} 
{LOTUS} uses a fixed window of nine consecutive Transformer blocks, $15$--$23$, for both models and all
datasets, motivated by prior evidence that hallucination- and factuality-related information is
prominent in intermediate Transformer representations
\citep{azaria2023internal,orgad2024llms,zhang2025icr}. We use ranks $(r_L,r_F)=(5,64)$ for $\bm C$ and
$\bm S$, and $(4,64)$ for $\bm V$, producing cores of dimensions $5\times64$, $5\times64$, and
$4\times64$. After vectorization and concatenation, each response is represented by an
$896$-dimensional feature vector. We then train a random-forest classifier with $200$ trees and
balanced class weights on these features and use its predicted hallucination probability for AUROC
evaluation. Layer window, Tucker ranks, and classifier settings are fixed across models and datasets
rather than tuned per benchmark. Appendix~\ref{app:window} evaluates alternative windows of the same
width and Appendix~\ref{app:ranks} the rank choices.

{\bf Baselines.}
We compare \textbf{LOTUS} against eight different hallucination detection baselines.

\textit{Label-free baselines.}
We consider six methods that require no hallucination-labeled training responses: answer-level
\emph{Perplexity}; LN-Entropy \citep{malinin2021uncertainty}, which aggregates predictive uncertainty
across generated responses; Semantic Entropy \citep{kuhn2023semantic}, over semantically equivalent
answer clusters; Lexical Similarity \citep{lin2022towards}, based on pairwise agreement among sampled
responses; EigenScore \citep{chen2024inside}, based on the eigenspectrum of hidden-state
representations; and HaloScope \citep{du2024haloscope}, which estimates a hallucination-related
subspace from unlabeled generations.

\textit{Supervised baselines.}
We additionally compare with two recent hidden-representation-based detectors trained on
hallucination labels. HARP \citep{hu2025harp} projects final-layer token representations onto a
low-dimensional subspace derived from the model's unembedding matrix and applies a lightweight MLP to
obtain token-level hallucination scores. ACT-ViT \citep{barshalom2025actvit} operates on layer--token
activation tensors, reducing them to a fixed resolution and training a ViT-based classifier for
response-level detection. Both are run with their released implementations and training settings;
further implementation details are given in Appendix~\ref{app:setup}. 
}
\vspace{-0.3cm}
{ 
\subsection{Experiment Results}
\label{sec:results}
\vspace{-0.3cm}

{\bf Comparison with baselines.} Table~\ref{tab:main} reports pooled AUROC under the question-level split across two LLMs and four QA
datasets. LOTUS outperforms HARP in all eight model--dataset combinations, by $2.15$ to $11.56$ AUROC
points and by $7.10$ points on average, with the largest gains on NQ-Open and TruthfulQA. Compared with ACT-ViT, LOTUS achieves the higher AUROC in six of the eight settings: by $0.18$,
$2.43$, and $5.04$ points on NQ-Open, TruthfulQA, and TyDiQA-GP with Qwen2.5-7B-Instruct, and by $2.38$
and $3.06$ points on NQ-Open and TyDiQA-GP with LLaMA-3.1-8B, where the two are effectively tied on
TruthfulQA ($89.19$ vs.~$89.07$). ACT-ViT is ahead on TriviaQA, by $1.08$ and $3.11$ points. TriviaQA is the largest of the
four datasets, at $99{,}600$ responses per model, and ACT-ViT retains a
layer--token activation tensor per response: on TriviaQA it stores $106$--$122$\,GB against
$28.6$--$32.7$\,GB for LOTUS at the token resolution we report, and about nineteen times the LOTUS
figure at its released resolution (Appendix~\ref{app:cost}). The remaining margin on that dataset is therefore obtained at a
considerably higher representation cost. LOTUS also outperforms every label-free baseline in all eight
settings. Overall, the proposed low-rank multilinear representation gives strong and consistent
detection performance across models and datasets at a small fraction of the representation size of the
strongest supervised baseline.

Table~\ref{tab:within_main} reports within-question AUROC under the question-level split. Label-free
baselines that assign a single score per question attain $50\%$ by construction and are omitted. LOTUS
outperforms HARP in all four settings by $5.40$--$14.95$ points and exceeds ACT-ViT in three of four,
trailing by $0.77$ points on TruthfulQA with LLaMA-3.1-8B. The larger gap between pooled and
within-question AUROC for HARP on TruthfulQA indicates that more of its pooled performance is
attributable to differences between questions, whereas LOTUS retains stronger response-level
discrimination. Standard deviations are larger than for pooled AUROC because the metric uses only
questions containing both correct and hallucinated responses; Appendix~\ref{app:within} reports the
number of eligible pairs.

}

{

\begin{table}[t]
\centering\small
\setlength{\tabcolsep}{4pt}
\caption{Pooled AUROC (\%) under the question-level split across two LLMs and four QA benchmarks. Results are averaged over five random trials; mean $\pm$ standard deviation is reported for supervised detectors. Best results are in bold.}
\label{tab:main}
\begin{tabular}{lcccc}
\toprule
\textbf{Method} & \textbf{NQ-Open} & \textbf{TruthfulQA} & \textbf{TriviaQA}
& \textbf{TyDiQA-GP} \\
\midrule
\midrule
\multicolumn{5}{l}{\textbf{Qwen2.5-7B-Instruct}} \\
Perplexity         & $57.58$ & $65.36$ & $75.42$ & $55.72$ \\
LN-Entropy         & $63.24$ & $59.61$ & $73.96$ & $68.60$ \\
Semantic Entropy   & $61.91$ & $60.86$ & $63.49$ & $51.60$ \\
Lexical Similarity & $65.95$ & $58.44$ & $71.09$ & $72.86$ \\
EigenScore         & $62.14$ & $56.24$ & $70.84$ & $63.48$ \\
HaloScope          & $60.49$ & $63.62$ & $74.19$ & $66.75$ \\
\midrule
HARP               & $67.73 \pm 2.24$ & $77.54 \pm 1.51$ & $88.75 \pm 0.33$ & $79.90 \pm 2.87$ \\
ACT-ViT            & $79.11 \pm 5.80$ & $85.40 \pm 1.01$ & $\mathbf{92.77 \pm 0.28}$ & $82.82 \pm 1.46$ \\
\textbf{LOTUS (Ours)} & $\mathbf{79.29 \pm 5.24}$ & $\mathbf{87.83 \pm 0.89}$ & $91.69 \pm 0.10$ & $\mathbf{87.86 \pm 1.95}$ \\
\midrule
\midrule
\multicolumn{5}{l}{\textbf{LLaMA-3.1-8B}} \\
Perplexity         & $41.37$ & $68.21$ & $51.13$ & $59.30$\\
LN-Entropy         & $66.25$ & $65.43$ & $54.82$ & $57.18$ \\
Semantic Entropy   & $54.21$ & $57.09$ & $63.43$ & $62.81$ \\
Lexical Similarity & $60.56$ & $58.73$ & $60.72$ & $55.15$ \\
EigenScore         & $71.34$ & $47.60$ & $62.24$ & $57.02$ \\
HaloScope          & $61.74$ & $69.37$ & $71.35$ & $51.52$ \\
\midrule
HARP               & $80.56 \pm 1.06$ & $79.58 \pm 5.38$ & $82.54 \pm 0.49$ & $74.99 \pm 0.88$ \\
ACT-ViT            & $82.98 \pm 1.76$ & $89.07 \pm 1.30$ & $\mathbf{87.80 \pm 0.33}$ & $79.43 \pm 1.41$ \\
\textbf{LOTUS (Ours)} & $\mathbf{85.36 \pm 0.42}$ & $\mathbf{89.19 \pm 1.41}$ & $84.69 \pm 0.20$ & $\mathbf{82.49 \pm 1.23}$ \\
\bottomrule
\bottomrule
\end{tabular}
\vspace{-0.5cm}
\end{table}

\begin{table}[t]
\centering\small
\setlength{\tabcolsep}{4pt}
\caption{
Within-question AUROC (\%) under the question-level split. Results are reported as mean $\pm$ standard deviation over five random trials. %
}
\label{tab:within_main}
\begin{tabular}{lcccc}
\toprule
& \multicolumn{2}{c}{\textbf{Qwen2.5-7B-Instruct}} & \multicolumn{2}{c}{\textbf{LLaMA-3.1-8B}} \\
\cmidrule(lr){2-3}\cmidrule(lr){4-5}
\textbf{Method} & \textbf{TruthfulQA} & \textbf{TyDiQA-GP} & \textbf{TruthfulQA} & \textbf{TyDiQA-GP} \\
\midrule
\midrule
HARP & $67.84 \pm 2.07$ & $82.69 \pm 1.44$ & $67.78 \pm 8.05$ & $75.87 \pm 1.63$ \\
ACT-ViT & $78.51 \pm 4.51$ & $81.93 \pm 1.93$ & $83.50 \pm 3.12$ & $80.50 \pm 2.34$ \\
\textbf{LOTUS (Ours)} & $\mathbf{82.07 \pm 2.18}$ & $\mathbf{88.09 \pm 3.77}$ & $\mathbf{82.73 \pm 1.92}$ & $\mathbf{81.34 \pm 3.01}$ \\
\midrule
\bottomrule
\end{tabular}
\end{table}

{
{\bf Different configurations of statistical summaries.}
Table~\ref{tab:configs-diff} evaluates how each token-mode summary defined in Section~\ref{sec:summaries}
contributes to LOTUS, which compresses and concatenates the peak activation ($\bm C$), activation
quantile ($\bm S$), and cross-layer difference ($\bm V$) summaries independently (see
\eqref{eq:features}). No single configuration is optimal in every model--dataset setting, but LOTUS is the most consistent.
It is the best configuration in two of the eight settings and stays within $0.74$ AUROC points of the
best alternative in the others. Removing individual summaries generally reduces performance, so
activation magnitude, distribution across tokens, and cross-layer variation carry complementary
signals. Projecting $\bm S$ and $\bm V$ with one shared pair of bases underperforms the default in
seven of eight settings, supporting separate bases per summary type.}

{
{\bf Cross-model transfer.}
LOTUS produces representations of the same dimensionality across LLMs, but in model-specific
coordinate systems, since the Tucker bases are learned independently from each model's activations. Therefore,
a detector trained on one model cannot directly interpret another model's LOTUS features even when
their dimensions match. To enable transfer, we learn a lightweight ridge mapping from the
target model's representation to the source model's on paired, unlabeled responses, and apply the
source detector to the aligned target representations without any target hallucination labels (see more details of the implementation in Appendix \ref{app:crossmodel}).
Table~\ref{tab:crossmodel_main} shows that the aligned detector transfers in both directions. Across
the four model--dataset settings, its pooled AUROC stays within $1.78$--$6.03$ points of a detector
trained directly on the target model, whereas removing the alignment causes substantial degradation,
particularly on TyDiQA-GP. %

}

\begin{table}[t]
\centering
\caption{
The full LOTUS configuration independently projects and concatenates
the peak activation ($\bm C$), activation quantile ($\bm S$), and cross-layer
difference ($\bm V$) summaries. The LOTUS row reports absolute pooled AUROC
(\%), while all ablated variants report the change in AUROC relative to
LOTUS ($\Delta$). Positive values indicate improvement over LOTUS and
negative values indicate degradation. Results use the question-level
split and are averaged over five random draws.}
\label{tab:configs-diff}
\resizebox{0.8\textwidth}{!}{%
\begin{tabular}{lcrrrr}
\toprule
& &  \multicolumn{4}{c}{\textbf{AUROC / $\Delta$ vs. LOTUS}} \\
\cmidrule(lr){3-6}
\textbf{Configuration} & \textbf{Summaries} 
& \textbf{NQ-Open} & \textbf{TruthfulQA}
& \textbf{TriviaQA} & \textbf{TyDiQA-GP} \\
\midrule

\multicolumn{6}{l}{\textit{Qwen2.5-7B-Instruct}}\\

\textbf{LOTUS (full)}
& $\bm C,\bm S,\bm V$ 
& $\mathbf{79.29}$ & $\mathbf{87.83}$
& $\mathbf{91.69}$ & $\mathbf{87.86}$ \\

\multicolumn{6}{l}{\hspace{0.5em}\textit{Ablated variants}}\\
w/o $\bm S$          & $\bm C,\bm V$  & $+0.15$ & $-0.04$ & $-0.23$ & $-0.06$ \\
Peak only        & $\bm C$    & $-0.48$ & $-0.75$ & $-0.41$ & $+0.40$ \\
Quantile only    & $\bm S$    & $-0.23$ & $-0.75$ & $-0.02$ & $-0.38$ \\
Difference only  & $\bm V$    & $-1.09$ & $-0.11$ & $-0.77$ & $-0.85$ \\
Shared bases     & $\bm S,\bm V$  & $-0.84$ & $-0.64$ & $+0.10$ & $-0.91$ \\

\midrule
\multicolumn{6}{l}{\textit{LLaMA-3.1-8B}}\\

\textbf{LOTUS (full)}
& $\bm C,\bm S,\bm V$ 
& $\mathbf{85.36}$ & $\mathbf{89.19}$
& $\mathbf{84.69}$ & $\mathbf{82.49}$ \\

\multicolumn{6}{l}{\hspace{0.5em}\textit{Ablated variants}}\\
w/o $\bm S$          & $\bm C,\bm V$  & $-0.18$ & $+0.13$ & $-0.63$ & $-1.04$ \\
Peak only        & $\bm C$    & $-0.30$ & $-0.60$ & $-0.49$ & $-0.82$ \\
Quantile only    & $\bm S$    & $-0.05$ & $-0.46$ & $+0.10$ & $+0.74$ \\
Difference only  & $\bm V$    & $-1.11$ & $+0.08$ & $-2.28$ & $-2.49$ \\
Shared bases     & $\bm S,\bm V$  & $-0.37$ & $-0.30$ & $-1.03$ & $-0.80$ \\

\bottomrule
\end{tabular}
}
\vspace{-0.7cm}
\end{table}

{\begin{table}[t]
\centering
\caption{
AUROC (\%) on the target model under the question-level split. \emph{Target-trained} uses target labels, while transferred detectors reuse the source classifier with or without unlabeled ridge alignment. Pooled AUROC is mean $\pm$ standard deviation over five draws; within-question AUROC is also reported.}
\label{tab:crossmodel_main}

\resizebox{0.8\columnwidth}{!}{%
\begin{tabular}{lcccc}
\toprule
& \multicolumn{2}{c}{\textbf{TruthfulQA}}
& \multicolumn{2}{c}{\textbf{TyDiQA-GP}} \\
\cmidrule(lr){2-3}\cmidrule(lr){4-5}
\textbf{Method}
& \textbf{Pooled} & \textbf{Within}
& \textbf{Pooled} & \textbf{Within} \\
\midrule

\multicolumn{5}{l}{\textit{Qwen2.5-7B-Instruct $\rightarrow$ LLaMA-3.1-8B}} \\
Target-trained LOTUS
& $89.22 \pm 1.44$ & $82.73$
& $82.47 \pm 1.24$ & $81.34$ \\

\textbf{Transfer + alignment}
& $\mathbf{87.44 \pm 1.18}$ & $\mathbf{80.67}$
& $\mathbf{77.26 \pm 2.32}$ & $\mathbf{78.44}$ \\

Transfer w/o alignment
& $63.73 \pm 4.93$ & $66.10$
& $48.79 \pm 2.31$ & $50.60$ \\

\midrule

\multicolumn{5}{l}{\textit{LLaMA-3.1-8B $\rightarrow$ Qwen2.5-7B-Instruct}} \\
Target-trained LOTUS
& $87.76 \pm 0.87$ & $82.07$
& $87.82 \pm 1.96$ & $88.09$ \\

\textbf{Transfer + alignment}
& $\mathbf{82.94 \pm 1.26}$ & $\mathbf{78.18}$
& $\mathbf{81.79 \pm 2.63}$ & $\mathbf{85.05}$ \\

Transfer w/o alignment
& $58.05 \pm 5.30$ & $59.86$
& $44.36 \pm 6.05$ & $43.42$ \\

\bottomrule
\end{tabular}%
}
\end{table}

}

{\bf Spectral evidence for low-rank hallucination structure.}
\label{sec:res-lowrank}
LOTUS relies on a low-dimensional feature basis, raising a natural question:
\emph{how much of the hidden-state variation is concentrated in a small number of directions?}
We analyze the peak-activation summary $\bm C_{i,k}$ one layer at a time. At layer $\ell$, each
response is represented by $\bm C_{i,k}[\ell,:]\in\mathbb{R}^{D}$, and separate covariance matrices
are formed for correct and hallucinated responses. To control for class imbalance, \emph{size-matched}
sampling uses an equal number of responses from each class; \emph{question-paired} sampling additionally
controls for question identity by selecting one correct and one hallucinated response from each question
containing both outcomes. Each scheme is repeated over $20$ draws. We characterize each covariance
spectrum by the variance captured by its leading $64$ eigen-directions and by the effective rank
$r_{\mathrm{eff}}=\exp(-\sum_j p_j\log p_j)$, where
$p_j=\lambda_j/\sum_m\lambda_m$.

Across the nine LOTUS layers, the leading $64$ of $D=3584$ directions explain $64.7$--$66.4\%$ of
the variance of correct responses and $52.3$--$55.5\%$ for hallucinated responses, indicating
substantial low-dimensional concentration. Hallucinated responses nevertheless exhibit broader spectra:
their effective rank is $2.01$--$2.22\times$ that of correct responses under size-matched sampling and
$1.52$--$1.60\times$ under question-paired sampling (Fig.~\ref{fig:spectrum}a,b). This ordering persists
across all layers, suggesting that hallucinated responses distribute their variance over a broader set
of feature directions even after controlling for question identity. Spectral separation alone, however,
does not determine the most discriminative layers. The embedding representation has the largest rank
ratio but only $76.5\%$ single-layer AUROC, whereas blocks $15$--$23$ average $86.8\%$
(Fig.~\ref{fig:spectrum}c). We therefore use this analysis to motivate low-rank compression while
selecting the layer window separately.

{ 

\begin{figure}[t]
\centering
\includegraphics[width=0.7\linewidth]{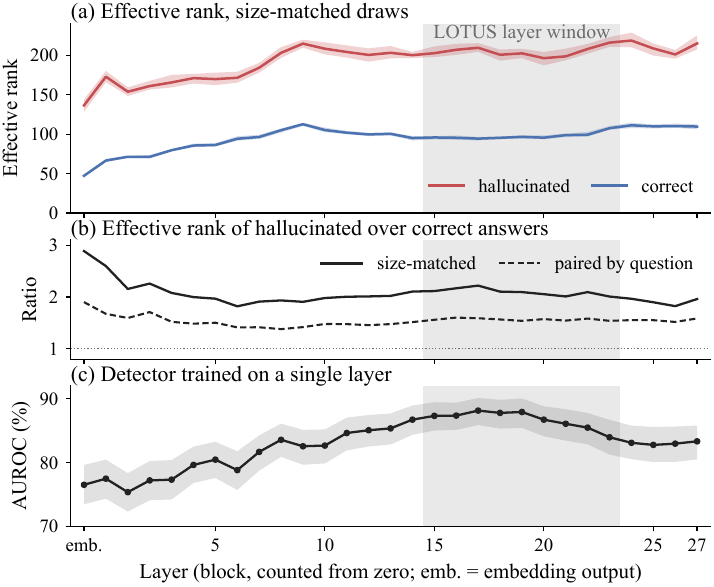}
\caption{Covariance spectra of the peak activation matrix for Qwen2.5-7B-Instruct on TyDiQA-GP, at every
layer. (a) effective rank of correct and hallucinated responses, (b) hallucinated-to-correct effective-rank ratio under size-matched and question-paired sampling, and (c) single-layer AUROC. Shaded bands indicate $95\%$ intervals; the gray region marks the LOTUS layer window.}
\label{fig:spectrum}
\vspace{-0.6cm}
\end{figure}

{\bf Performance with limited labeled data.} We further examine whether the compact Tucker-2 representation helps when fewer labeled training
questions are available. Table~\ref{tab:lessdata} in Appendix~\ref{app:lessdata} reports the
AUROC difference between Tucker-2 and an equal-dimensional flat-PCA representation as the training set
is reduced to $5\%$--$50\%$, with the test set held fixed. With the full training set, the two
representations perform similarly, within $0.80$ AUROC points in all settings. As the training data
decrease, Tucker-2 gains a clear advantage in several settings: on LLaMA-3.1-8B with TruthfulQA it
improves AUROC by $2.66$--$3.98$ points in every reduced-data regime, and on Qwen2.5-7B-Instruct with
TyDiQA-GP the gain reaches $5.06$ points. The remaining settings are mixed, so the benefit is not universal, but the more compact, structured
Tucker-2 parameterization is advantageous when labeled training data are scarce.

{\bf Tucker-3 token-mode extension.}
LOTUS aggregates the token mode before decomposition, trading explicit token-order information for a
compact layer--feature representation. To examine this trade-off, we construct a Tucker-3 variant that
partitions each response into $B$ contiguous token bins and additionally decomposes the resulting
token mode with rank $r_T$. Table~\ref{tab:tucker3} in Appendix~\ref{app:reductions} shows that
retaining this coarse temporal structure provides only limited benefit. With the proposed
order-statistic summaries, no Tucker-3 configuration improves pooled AUROC over the Tucker-2 baseline
by more than $0.37$ points across the four settings, while increasing the representation from $896$ to
as many as $7{,}168$ features, and increasing the token rank gives no consistent gain. The same trend holds with average pooling, supporting the choice to aggregate the token mode before
decomposition, i.e., the Tucker-2 representation is substantially more compact at comparable detection
performance.

{

 \section{Conclusion}

In  this work, we introduced \textbf{LOTUS}, a low-rank multilinear framework for hallucination detection from LLM hidden representations. \textbf{LOTUS} converts variable-length token representations into complementary layer--feature summaries and learns compact Tucker-2 subspaces from unlabeled activations, requiring hallucination labels only for a lightweight downstream classifier. Across two LLMs and four question-answering benchmarks, our method achieves strong and consistent detection performance while substantially reducing representation size and supporting transfer across models with different internal dimensions. Extensive ablations further validate the contributions of the proposed summaries, low-rank projection, rank choices, and transfer design. Overall, our results show that structured low-dimensional representations provide an effective and efficient basis for detecting hallucinations from LLM internal states.

\subsubsection*{AI use statement}

We used a generative AI assistant in two ways. First, for code: most of the pipeline, job and analysis
scripts in the released repository were written or revised with AI assistance under the authors'
direction. Every number we report is produced by that code, the scripts carry self-tests that check
them against independent implementations where one exists (the ACT-ViT preprocessing, for instance, is
asserted bit-identical to the released implementation of its authors), and the authors reviewed both
the code and its outputs. Second, for writing: the assistant was used to draft and revise text
throughout the paper, and to locate candidate related work, which the authors then verified against
the original papers. The assistant was not used to generate experimental results, data or labels: all
generations, activations, metrics and tables come from running the released code. We have reviewed all
AI-assisted work and take responsibility for the final content of this paper, including its text,
claims and artifacts.

\subsubsection*{Ethics statement}

This work uses four public question answering benchmarks and two openly available language models,
each under its own license. It involves no human subjects, no new data collection, and no personally
identifying information beyond what those public datasets already contain. Hallucination labels are
assigned by automatic metrics rather than by human annotation (Section~\ref{sec:setup}), so they inherit the
biases of those metrics; we therefore report their agreement with an independently released judge in
Appendix~\ref{app:labels} rather than treating them as ground truth. A detector of this kind reduces,
but does not remove, the risk of acting on an incorrect model output, and accuracy on these benchmarks
does not imply reliability in deployment, in other languages, or on other domains; we make no such
claim. We are not aware of dual-use concerns raised by this work beyond those already present in the
publicly released models it analyses.

\subsubsection*{Reproducibility statement}

The detector is specified in Section~\ref{sec:method}: the token summaries, the entry-wise scaling, the two
Tucker-2 bases and the classifier. The configuration behind every reported number is given in
Section~\ref{sec:setup}, which states the layer window, the ranks, the classifier, the split and the two
metrics, and Appendix~\ref{app:setup} adds the generation settings, the prompts, the per-question
seeding and how the two trained baselines were run. A single harness owns the splits, the seeds and
the metrics so that no method computes its own, and Appendix~\ref{app:split} defines both split
procedures. Code for generation, extraction, evaluation, the ablations and the analysis is available
at \url{https://github.com/devvansh1997/LOTUS}, one script per stage, each with a self-test. The generated
responses and activations are not released because of their size; each question is seeded from a
global seed and its identifier, so they can be regenerated exactly from the released configuration.

\bibliography{iclr2027_conference}  %
\bibliographystyle{iclr2027_conference}

\clearpage

\appendix

\begin{center}
    {\bf Supplementary Material of ``LLM Hallucination Detection via Low-rank Multilinear Decomposition''}
\end{center}

\section{Related Work}
\label{sec:related}

\paragraph{Hallucination detection in LLMs.} Hallucination refers to unfaithful, unsupported, or
fabricated generations produced by large language models \citep{ji2023survey,huang2025survey}. Since
hallucinations reduce the reliability of LLMs in downstream applications, hallucination detection has
become an important research direction. Existing methods can be broadly grouped into uncertainty-based
approaches \citep{chen2025enhancing,liu2026noise} and internal-representation-based approaches
\citep{zhang2025icr}. Uncertainty-based methods assume that hallucinated outputs are associated with
lower model confidence or higher variability across generations. Early work measures token-level
uncertainty using predictive logits, entropy, or likelihood-based scores such as perplexity
\citep{huang2025reppl}. Other studies extend this idea to sentence-level or response-level uncertainty
\citep{malinin2021uncertainty} by aggregating token probabilities or comparing multiple sampled
generations. For example, semantic entropy \citep{farquhar2024detecting} estimates the consistency of
meanings across multiple outputs, while lexical similarity  \citep{lin2022towards}
measures whether sampled answers agree with each other at the surface text level.

\paragraph{Internal representation and probing methods.} To overcome the limitations of output-only
detection, recent work investigates whether internal representations of LLMs encode
truthfulness-related information \citep{ji2024llm,orgad2024llms}. Probing-based approaches
\citep{alain2016understanding,ju2024how,orgad2024llms} train classifiers on hidden states, logits, or
activation vectors to predict whether a generated response is correct or hallucinated. These methods
are motivated by the observation that LLMs may internally encode more knowledge than they explicitly
express in the final output \citep{duan2024llms}. Several works \citep{kim2025detecting,barshalom2025actvit}
analyze layer-wise or token-wise representations and show that hallucination signals can emerge in
intermediate hidden states. HaloScope \citep{du2024haloscope}, for instance, leverages unlabeled
embeddings and applies SVD to identify important subspace directions before probing them for
hallucination detection. EigenScore \citep{chen2024inside} measures semantic agreement through
covariance eigenvalues.

\paragraph{Mechanistic interpretability of LLMs.} Mechanistic interpretability provides another
perspective for understanding hallucinations by analyzing model parameters, hidden states, and
layer-wise computation \citep{merullo2024talking,cheng2024enhancing}. Prior work has studied weight
matrices \citep{hu2025harp}, attention heads \citep{sriramanan2024llmcheck}, and feed-forward modules
\citep{zhang2025icr} to uncover structural properties of LLMs. In reasoning models, hallucinations may
arise from shallow pattern matching, unstable reasoning trajectories, incorrect backtracking, or
overthinking \citep{wang2026joint}. These findings suggest that hallucination is not only a
surface-level text generation problem, but also a consequence of abnormal internal computation.

\section{LOTUS workflow} 
Algorithms~\ref{alg:fit} and~\ref{alg:score} summarize the training and inference procedures of LOTUS. During training, the scaling statistics and the Tucker-2 bases $(\bm U_L, \bm U_F)$ are estimated separately for each summary type from the training responses without hallucination labels, which are used only to train the final classifier. At inference, a test response is scaled with the training statistics and projected onto the same fixed bases, so scoring requires two matrix multiplications per summary followed by a single classifier evaluation.

\begin{figure}[h]
\begin{minipage}{\linewidth}
\begin{algorithm}[H]
\caption{Training the proposed detector}
\label{alg:fit}
\begin{algorithmic}[1]
\REQUIRE Training responses $ \{ \bm a_{i,k}\}_{i \in N_{\text{train}}, k=1}^{K}$, labels $z_{i,k}$, layer window $\mathcal{W}$
\FOR{each response $\bm a_{i,k}$}
    \STATE Compute summaries $\bm C_{i,k}, \bm S_{i,k}, \bm V_{i,k}$
\ENDFOR
\FOR{each summary $\tilde{\bm X}_{i,k} \in \{\bm C_{i,k}, \bm S_{i,k}, \bm V_{i,k}\}$}
    \STATE Estimate robust scaling statistics from $\tilde{\bm X}_{i,k}$, without labels
    \STATE Fit layer basis $\bm U_L$ and feature basis $\bm U_F$ with truncated HOSVD via (\ref{eq:tucker-objective})
\ENDFOR
\FOR{each response $\bm a_{i,k}$}
    \STATE Compute core $\bm G_{i,k}=\bm U_L^{\top}\tilde{\bm X}_{i,k}\bm U_F$ for each summary $\tilde{\bm X}_{i,k}$ with (\ref{eq:core})
    \STATE Concatenate cores into feature vector with (\ref{eq:features})
    \[
    \boldsymbol{\phi}_{i,k}= \left(\mathrm{vec}(\bm G^C_{i,k})^\top, \mathrm{vec}(\bm G^S_{i,k})^\top, \mathrm{vec}(\bm G^V_{i,k})^\top\right)^\top
    \]
\ENDFOR
\STATE Train classifier on $\{(\boldsymbol{\phi}_{i,k}, z_{i,k})\}$
\end{algorithmic}
\end{algorithm}
\end{minipage}

\vspace{0.5em}

\begin{minipage}{\linewidth}
\begin{algorithm}[H]
\caption{Scoring a test response}
\label{alg:score}
\begin{algorithmic}[1]
\REQUIRE Test response $ \{ \bm a_{i,k}\}_{i \in N_{\text{test}}, k=1}^{K}$, fitted scaling statistics, bases $(\bm U_L, \bm U_F)$, trained classifier
\STATE Compute summaries $\bm C_{i,k}, \bm S_{i,k}, \bm V_{i,k}$ for each response $\bm a_{i,k}$
\STATE Scale each summary with the training statistics
\STATE Project each scaled summary onto $(\bm U_L, \bm U_F)$ to obtain its core $\bm G_{i,k}$ via (\ref{eq:core})
\STATE Concatenate cores into feature vector $\boldsymbol{\phi}_{i,k}$ using (\ref{eq:features})
\STATE \textbf{return} $c (\bm a_{i,k}) \gets$ classifier's predicted probability with $\boldsymbol{\phi}_{i,k}$
\end{algorithmic}
\end{algorithm}
\end{minipage}
\end{figure}

\section{Layer window analysis}
\label{app:window}

LOTUS uses residual-stream representations from blocks 15–23 for both models. To examine the sensitivity of different layer windows, we evaluate the hallucination signal available at each individual layer using Qwen2.5-7B-Instruct on TyDiQA-GP and TruthfulQA shown in Table~\ref{tab:window}. Across both datasets, performance is lower for the earliest windows, the intermediate-layer windows yield consistently strong AUROC with only small variations across neighboring windows, and performance generally declines as the window moves toward the final layers. The selected nine-block window obtains the best intermediate-layer AUROC performance, indicating a strong hallucination-discriminative signal.

Table~\ref{tab:layers} reports the AUROC of a detector trained on a single layer for Qwen2.5-7B-Instruct. On both datasets, performance is lowest near the embedding layer, rises through the early and middle blocks, and peaks within our window (blocks 15--23), reaching $89.0\%$ on TyDiQA-GP and $87.9\%$ on TruthfulQA at blocks 16--18. Beyond the window, AUROC declines toward the final block. The range summary matches or exceeds the peak summary in almost every block, indicating that activation distributions across tokens carry additional signal beyond the maximum.

\begin{table}[h]
\centering\small
\setlength{\tabcolsep}{4pt}
\caption{ Pooled AUROC (\%) under the question-level split, averaged over five different runs. Results are reported across different nine-layer windows while keeping all other settings fixed.}
\label{tab:window}
\begin{tabular}{lcc}
\toprule
\textbf{Window (blocks)} & \textbf{TruthfulQA} & \textbf{TyDiQA-GP} \\
\midrule
\multicolumn{3}{l}{\textit{Qwen2.5-7B-Instruct}} \\
$3$--$11$ & $87.17$ & $84.93$ \\
$7$--$15$ & $87.13$ & $87.04$ \\
$11$--$19$ & $87.52$ & $87.27$ \\
$13$--$21$ & $87.92$ & $87.48$ \\
$15$--$23$ (reported) & $87.50$ & $87.89$ \\
$19$--$27$ & $85.77$ & $86.55$ \\
\midrule
\multicolumn{3}{l}{\textit{LLaMA-3.1-8B}} \\
$3$--$11$ & $91.57$ & $83.98$ \\
$7$--$15$ & $91.17$ & $85.64$ \\
$11$--$19$ & $90.83$ & $84.74$ \\
$15$--$23$ (reported) & $89.35$ & $82.46$ \\
$17$--$25$ & $88.52$ & $80.51$ \\
$18$--$26$ & $88.38$ & $80.29$ \\
$19$--$27$ & $88.27$ & $79.79$ \\
$23$--$31$ & $87.84$ & $79.49$ \\
\bottomrule
\end{tabular}
\end{table}

\begin{table}[h]
\centering\small
\setlength{\tabcolsep}{4pt}
\caption{Single-layer AUROC (\%) for Qwen2.5-7B-Instruct: the peak or range summary of one layer, projected to $64$ features and classified by logistic regression, with question-grouped five-fold cross-validation. Blocks are counted from zero; the rules enclose our window, blocks $15$ to $23$.}
\label{tab:layers}
\begin{tabular}{lcccc}
\toprule
& \multicolumn{2}{c}{\textbf{TyDiQA-GP}} & \multicolumn{2}{c}{\textbf{TruthfulQA}} \\
\cmidrule(lr){2-3}\cmidrule(lr){4-5}
\textbf{Block} & Peak & Range & Peak & Range \\
\midrule
Embedding & 76.5 & 77.3 & 82.3 & 82.3 \\
0         & 77.5 & 78.1 & 83.9 & 85.0 \\
1         & 75.3 & 79.5 & 83.7 & 84.9 \\
2         & 77.2 & 78.9 & 85.4 & 86.3 \\
3         & 77.3 & 80.0 & 85.8 & 86.5 \\
4         & 79.6 & 82.3 & 86.2 & 86.4 \\
5         & 80.5 & 83.2 & 86.4 & 87.3 \\
6         & 78.8 & 83.0 & 86.2 & 87.5 \\
7         & 81.7 & 84.6 & 85.9 & 86.7 \\
8         & 83.6 & 85.7 & 87.1 & 87.5 \\
9         & 82.6 & 84.0 & 85.2 & 85.8 \\
10        & 82.7 & 84.9 & 85.6 & 86.4 \\
11        & 84.6 & 85.9 & 85.8 & 86.7 \\
12        & 85.1 & 85.9 & 85.7 & 86.8 \\
13        & 85.4 & 86.7 & 85.5 & 86.0 \\
14        & 86.7 & 87.1 & 86.4 & 87.1 \\
\midrule
15        & 87.3 & 87.8 & 86.6 & 87.5 \\
16        & 87.4 & 88.2 & 87.4 & 87.9 \\
17        & 88.2 & 89.0 & 87.5 & 87.9 \\
18        & 87.8 & 88.6 & 87.4 & 87.9 \\
19        & 87.9 & 87.9 & 87.2 & 87.8 \\
20        & 86.7 & 87.1 & 87.0 & 87.8 \\
21        & 86.1 & 86.6 & 86.4 & 87.2 \\
22        & 85.5 & 85.4 & 86.1 & 86.3 \\
23        & 84.0 & 85.0 & 84.8 & 84.9 \\
\midrule
24        & 83.1 & 84.1 & 84.8 & 85.0 \\
25        & 82.8 & 83.9 & 85.1 & 85.7 \\
26        & 83.0 & 83.6 & 85.2 & 85.6 \\
27        & 83.3 & 83.8 & 83.4 & 83.8 \\
\bottomrule
\end{tabular}
\end{table}
\section{Other implementation details}
\label{app:setup}

We conducted all experiments for all methods on a single NVIDIA H100 80GB GPU.

\paragraph{Generation.}
For each question, we generate $K=10$ answers using beam search with sampling: $10$ beams, $10$ returned sequences, temperature $0.5$, top-$k=5$, top-$p=0.99$, and at most $64$ new tokens. Prompts use plain text without a chat template; TyDiQA-GP additionally includes the passage and an instruction to answer from it. Each question uses a fixed seed for reproducibility, duplicate TriviaQA questions are merged, and the same generated responses are used across all methods.

\paragraph{HARP.} We run the released implementation on our answers and labels. HARP projects final-layer token hidden states onto the $256$ smallest-singular-value directions of the unembedding matrix, applies a two-layer MLP, and uses the maximum token score as the answer score. We report the performance under both question-level and answer-level splits in Table~\ref{tab:std-actvit} (Appendix~\ref{app:std}).

\paragraph{ACT-ViT.}
We follow the released ACT-ViT implementation and same training setup. We train for $15$ epochs with a batch size of $128$, learning rate and weight decay of $10^{-3}$, using a three-layer transformer with four attention heads, hidden size $128$, and dropout $0.3$. Activations are max-pooled using the released preprocessing to $8$ layers and $N=20$ token positions. We use $N=20$ instead of the default $N=100$ because our generated responses are short and larger values would mostly introduce padding. However, we also report results with $N=100$ in Table \ref{tab:neff} (Appendix~\ref{app:neff}). %

\paragraph{Evaluation.}
For each seed in $\{42,0,1,2,3\}$, all methods use the same train/test questions and compute the metrics from the scores the method
returns. This ensures that all methods use identical data partitions and evaluation procedures. 

\paragraph{Summary statistics scaling in LOTUS.} 
Before doing Tucker-2 decomposition in LOTUS, each summary is scaled entry by entry using statistics computed from the
training responses alone: every entry is clipped to its $0.5$th and $99.5$th training
percentiles, centered by its median, divided by its interquartile range rescaled to estimate a
standard deviation, and clipped to $[-6,6]$. We use the median and the interquartile range
rather than the mean and the standard deviation because a few features of the residual stream
take values far larger than the rest \citep{sun2024massive} and would otherwise dominate the
objective in \eqref{eq:tucker-objective}. We write $\tilde{\bm X}_{i,k}$ for the scaled summary.

\section{Effect of the data split}
\label{app:split}
\label{sec:res-leak}
As discussed in Section~\ref{sec:splits}, the answer-level split allows responses from the same question to appear in both training and test sets. This appendix quantifies the resulting AUROC inflation across trained detectors and explains why this overlap is not apparent from aggregate results. Figure~\ref{fig:splits} illustrates the two splitting strategies. 

\begin{figure}[t]
\centering
\includegraphics[width=\linewidth]{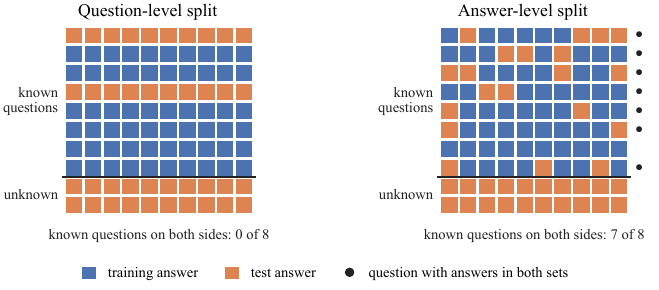}
\caption{Comparison of question-level and answer-level splitting with ten responses per question. Question-level splitting keeps all responses from a known question in one partition, whereas answer-level splitting can place responses from the same question in both training and test sets, affecting $92.6\%$--$95.7\%$ of known questions in our data.}
\label{fig:splits}
\end{figure}

\label{sec:leak}
\paragraph{Overlap between training and test.} A known question stays out of the test set only if all
ten of its answers land in training, and out of training only if all ten land in test. Under the answer-level split, a question appears in both training and test sets with probability
$1-0.75^{10}-0.25^{10}\approx0.944$. Empirically, this overlap ranges from $92.6\%$ to $95.7\%$ across the two models, TruthfulQA and TyDiQA-GP, over five runs. The test answers are new, but their questions were seen in
training, and answers to the same question share a prompt and a topic. A detector may achieve high performance by recognizing question-specific information rather than evaluating the factuality of the response itself.

\label{sec:invisible}
\paragraph{Impact of the splitting protocol.}
The question- and answer-level splits produce nearly identical train/test sizes, making the split choice difficult to infer from aggregate results alone. The released HARP implementation splits answers, and ACT-ViT splits responses in a similar way. When multiple responses belong to the same question, this can place the same question in both the training and test sets. To isolate the effect of the split, we evaluate each trained detector under both settings while keeping the answers, labels, features, and training procedure fixed as shown in Table \ref{tab:leak}. Across HARP, LOTUS, and both ACT-ViT variants, the answer-level split consistently yields higher AUROC, with gains ranging from less than one point to over $13$ points. Under this split, LOTUS achieves the highest AUROC in hallucination detection performance.

\begin{table}[h]
\centering\small
\setlength{\tabcolsep}{3.5pt}
\caption{Pooled AUROC (\%) of HARP and LOTUS under the answer-level split (Ans.), and its
gain over the question-level split ($\Delta$ = Ans. $-$ question-level AUROC in
Table~\ref{tab:main}). ACT-ViT under both splits is in Table~\ref{tab:std-actvit}. Standard
deviations are in Table~\ref{tab:std-main}.}
\label{tab:leak}
\begin{tabular}{lrrrrrrrr}
\toprule
& \multicolumn{2}{c}{\textbf{NQ-Open}} & \multicolumn{2}{c}{\textbf{TruthfulQA}}
& \multicolumn{2}{c}{\textbf{TriviaQA}} & \multicolumn{2}{c}{\textbf{TyDiQA-GP}} \\
\cmidrule(lr){2-3}\cmidrule(lr){4-5}\cmidrule(lr){6-7}\cmidrule(lr){8-9}
\textbf{Method} & Ans. & $\Delta$ & Ans. & $\Delta$ & Ans. & $\Delta$ & Ans. & $\Delta$ \\
\midrule
\multicolumn{9}{l}{\textit{Qwen2.5-7B-Instruct}} \\
HARP & 81.03 & $+13.30$ & 85.16 & $+7.62$ & 92.50 & $+3.75$ & 91.75 & $+11.85$ \\
LOTUS & 91.14 & $+11.85$ & 94.03 & $+6.20$ & 94.05 & $+2.36$ & 94.72 & $+6.86$ \\
\midrule
\multicolumn{9}{l}{\textit{LLaMA-3.1-8B}} \\
HARP & 86.56 & $+6.00$  & 84.84 & $+5.26$ & 88.08 & $+5.54$ & 85.65 & $+10.66$ \\
LOTUS & 91.96 & $+6.60$  & 94.41 & $+5.22$ & 89.48 & $+4.79$ & 90.07 & $+7.58$ \\
\bottomrule
\end{tabular}
\end{table}
\section{Comparison with Alternative Reductions}
\label{app:reductions}

\paragraph{Reductions of equal width.} 

Table~\ref{tab:controls} compares Tucker-2 with Tucker-3, flat PCA, and random projection, all using the same random-forest classifier (Section~\ref{sec:setup}). Tucker-2 matches flat PCA and Tucker-3 in accuracy while using far fewer parameters and multiply-adds, and random projection performs clearly worse. The same pattern holds under within-question AUROC (defined in Section~\ref{sec:reporting}). %

\begin{table}[t]
\centering\small
\setlength{\tabcolsep}{3pt}
\caption{ Performance comparison of reduction methods using the same random-forest classifier under the question-level split, averaged over five runs on TruthfulQA (TQA) and TyDiQA-GP (TyDi). We report pooled AUROC (\%), feature dimension, fitted basis parameters, and projection multiply-adds for Qwen2.5-7B-Instruct / LLaMA-3.1-8B.  %
}
\label{tab:controls}
\begin{tabular}{lccccccc}
\toprule
& & & & \multicolumn{2}{c}{\textbf{Qwen2.5-7B}} & \multicolumn{2}{c}{\textbf{LLaMA-3.1-8B}} \\
\cmidrule(lr){5-6}\cmidrule(lr){7-8}
\textbf{Reduction} & \textbf{Features} & \textbf{Parameters} & \textbf{Multiply-adds} & TQA & TyDi & TQA & TyDi \\
\midrule
\textbf{Tucker-2 (LOTUS)}    & $896$     & $1.15$ / $1.31$ & $9.9$ / $11.3$  & 87.76 & 87.82 & 89.22 & 82.47 \\
Tucker-3, 2 bins, $r_T{=}1$ & $896$     & $1.15$ / $1.31$ & $19.7$ / $22.6$ & 87.86 & 87.96 & 89.51 & 82.67 \\
Tucker-3, 8 bins, $r_T{=}8$ & $7{,}168$ & $1.15$ / $1.31$ & $79.1$ / $90.3$ & 86.84 & 87.28 & 89.20 & 80.96 \\
Flat PCA                    & $896$     & $45.6$ / $52.2$ & $45.6$ / $52.2$ & 88.09 & 87.45 & 88.42 & 81.96 \\
Random projection           & $896$     & none            & $45.6$ / $52.2$ & 82.87 & 85.38 & 84.32 & 77.24 \\
\bottomrule
\end{tabular}
\end{table}

\paragraph{Single-layer reductions.} Table~\ref{tab:controls-extra} compares Tucker-2 with PCA applied either to the best single layer or to the average over the selected layer window, using the same $896$ features. Their AUROC remains close to Tucker-2, ranging from $1.55$ points lower to $0.65$ points higher. However, both alternatives require about $4.6\times$ more fitted parameters, showing that Tucker-2 achieves comparable performance with a substantially more compact representation.

\begin{table}[h]
\centering\small
\setlength{\tabcolsep}{4pt}
\caption{Single-layer reductions to $896$ features with the random forest, under the question-level
split: pooled AUROC (\%), averaged over five draws. Parameters are the entries of the fitted bases, for
Qwen2.5-7B-Instruct / LLaMA-3.1-8B.}
\label{tab:controls-extra}
\begin{tabular}{lccccc}
\toprule
& & \multicolumn{2}{c}{\textbf{Qwen2.5-7B-Instruct}} & \multicolumn{2}{c}{\textbf{LLaMA-3.1-8B}} \\
\cmidrule(lr){3-4}\cmidrule(lr){5-6}
\textbf{Reduction} & \textbf{Parameters} & TruthfulQA & TyDiQA-GP & TruthfulQA & TyDiQA-GP \\
\midrule
\textbf{Tucker-2 (LOTUS)} & 1.15M / 1.31M & 87.76 & 87.82 & 89.22 & 82.47 \\
Best layer               & 5.28M / 6.03M & 88.05 & 86.54 & 89.11 & 83.12 \\
Layer average            & 5.28M / 6.03M & 87.87 & 86.97 & 88.28 & 80.92 \\
\bottomrule
\end{tabular}
\end{table}
\paragraph{Tucker-3.}
We extend Tucker-2 by dividing each response into $B$ token bins and applying truncated HOSVD over the layer, token, and feature modes. Within each bin, we use either the three order-statistic summaries from \S\ref{sec:summaries} or average pooling over hidden states and layer updates. We evaluate $B\in\{1,2,4,8\}$ with corresponding token ranks $r_T\in\{B/2,B\}$, while keeping the layer and feature ranks fixed. All projection bases are fitted on training responses without labels, and the resulting cores are passed to the same random-forest classifier. The one-bin order-statistic setting corresponds to the Tucker-2 baseline. As shown in Table~\ref{tab:tucker3}, retaining an explicit token mode yields limited improvement. No order-statistic setting improves over one bin by more than $0.37\%$ in AUROC, while full token rank is usually worse. Average pooling shows a similar pattern, suggesting that preserving token structure explicitly does not consistently improve detection performance.

\begin{table}[h]
\centering\small
\setlength{\tabcolsep}{3.5pt}
\caption{Tucker-3 with $B$ token bins and token rank $r_T$, with the random forest, under the
question-level split: pooled AUROC (\%), averaged over five draws, and in parentheses the difference from
one bin of the same pooling over the same draws. Features are passed to the classifier.}
\label{tab:tucker3}
\begin{tabular}{rrrcccc}
\toprule
& & & \multicolumn{2}{c}{\textbf{Qwen2.5-7B-Instruct}} & \multicolumn{2}{c}{\textbf{LLaMA-3.1-8B}} \\
\cmidrule(lr){4-5}\cmidrule(lr){6-7}
$B$ & $r_T$ & \textbf{Features} & TruthfulQA & TyDiQA-GP & TruthfulQA & TyDiQA-GP \\
\midrule
\multicolumn{7}{l}{\textit{Order statistics within bins (peak, range, update)}} \\
$1$ & $1$ & $896$ & $87.49$ & $87.88$ & $89.34$ & $82.46$ \\
$2$ & $1$ & $896$ & $87.86$ ($+0.37$) & $87.96$ ($+0.08$) & $89.51$ ($+0.17$) & $82.67$ ($+0.21$) \\
$2$ & $2$ & $1{,}792$ & $87.30$ ($-0.19$) & $87.79$ ($-0.10$) & $89.38$ ($+0.03$) & $81.81$ ($-0.65$) \\
$4$ & $2$ & $1{,}792$ & $87.50$ ($+0.01$) & $88.07$ ($+0.18$) & $89.55$ ($+0.20$) & $81.82$ ($-0.65$) \\
$4$ & $4$ & $3{,}584$ & $87.22$ ($-0.27$) & $87.61$ ($-0.27$) & $89.18$ ($-0.17$) & $81.57$ ($-0.89$) \\
$8$ & $4$ & $3{,}584$ & $86.97$ ($-0.52$) & $87.46$ ($-0.43$) & $89.55$ ($+0.21$) & $81.66$ ($-0.81$) \\
$8$ & $8$ & $7{,}168$ & $86.84$ ($-0.65$) & $87.28$ ($-0.61$) & $89.20$ ($-0.14$) & $80.96$ ($-1.51$) \\
\midrule
\multicolumn{7}{l}{\textit{Average pooling within bins (state, update)}} \\
$1$ & $1$ & $576$ & $87.51$ & $86.94$ & $89.51$ & $81.87$ \\
$2$ & $1$ & $576$ & $87.44$ ($-0.07$) & $86.84$ ($-0.09$) & $89.70$ ($+0.19$) & $82.40$ ($+0.52$) \\
$2$ & $2$ & $1{,}152$ & $87.21$ ($-0.30$) & $87.17$ ($+0.23$) & $89.79$ ($+0.28$) & $81.71$ ($-0.17$) \\
$4$ & $2$ & $1{,}152$ & $86.86$ ($-0.65$) & $87.47$ ($+0.53$) & $89.66$ ($+0.15$) & $82.06$ ($+0.19$) \\
$4$ & $4$ & $2{,}304$ & $86.66$ ($-0.84$) & $87.17$ ($+0.24$) & $89.22$ ($-0.28$) & $81.66$ ($-0.21$) \\
$8$ & $4$ & $2{,}304$ & $86.65$ ($-0.85$) & $87.33$ ($+0.39$) & $89.85$ ($+0.35$) & $81.23$ ($-0.64$) \\
$8$ & $8$ & $4{,}608$ & $86.78$ ($-0.72$) & $87.14$ ($+0.20$) & $89.09$ ($-0.42$) & $80.58$ ($-1.30$) \\
\bottomrule
\end{tabular}
\end{table}

\section{Performance with Limited Training Data}
\label{app:lessdata}

We further compare the Tucker-2 core with flat PCA under reduced training data, using only $5\%$--$50\%$ of the training questions while keeping the test sets fixed (Table~\ref{tab:lessdata}). The Tucker-2 core shows a clear advantage when training data are limited. In the other settings, its relative performance varies with the amount of available training data. Tucker-2 improves AUROC by up to $5.06\%$  on Qwen2.5-7B-Instruct with TyDiQA-GP and by up to $3.98\%$  on LLaMA-3.1-8B with TruthfulQA using only $5\%$--$10\%$ of the training data. At very small training sizes, flat PCA may not produce the full $896$ features because the number of principal components is limited by the number of training samples. These results suggest that the more compact Tucker-2 parameterization shows stronger robustness when training data is limited.

\begin{table}[h]
\centering\small
\setlength{\tabcolsep}{5pt}
\caption{Reducing the number of labeled training questions, with the test sets held fixed and both
reductions refit on each run. For each setting the first row is the pooled AUROC (\%) of the Tucker-2
core, averaged over five runs, and the second is the difference of flat PCA of the same width from it,
in AUROC points; negative values are below the core. The last row averages the four differences.
$^\dagger$At these sizes flat PCA cannot supply all $896$ features, since each summary yields at most
one component per training response.}
\label{tab:lessdata}
\begin{tabular}{llccccc}
\toprule
\textbf{Setting} & & \textbf{$5\%$} & \textbf{$10\%$} & \textbf{$25\%$} & \textbf{$50\%$} & \textbf{$100\%$} \\
\midrule
\multicolumn{7}{l}{\textit{Qwen2.5-7B-Instruct}} \\
TruthfulQA & Tucker-2 & $77.10$ & $81.32$ & $84.89$ & $86.12$ & $87.76$ \\
           & flat PCA & $+0.09^\dagger$ & $+0.64$ & $-1.85$ & $-0.71$ & $+0.33$ \\
TyDiQA-GP  & Tucker-2 & $74.05$ & $84.00$ & $85.01$ & $86.31$ & $87.82$ \\
           & flat PCA & $-0.18^\dagger$ & $-5.06^\dagger$ & $-1.33$ & $-1.57$ & $-0.38$ \\
\midrule
\multicolumn{7}{l}{\textit{LLaMA-3.1-8B}} \\
TruthfulQA & Tucker-2 & $81.60$ & $79.51$ & $84.02$ & $87.49$ & $89.22$ \\
           & flat PCA & $-3.98^\dagger$ & $-2.70$ & $-2.91$ & $-2.66$ & $-0.80$ \\
TyDiQA-GP  & Tucker-2 & $68.48$ & $73.82$ & $77.38$ & $79.84$ & $82.47$ \\
           & flat PCA & $+0.44^\dagger$ & $-3.84^\dagger$ & $+1.23$ & $+0.03$ & $-0.51$ \\
\cmidrule(lr){3-7}
\textit{average} & flat PCA & $-0.91$ & $-2.74$ & $-1.21$ & $-1.23$ & $-0.34$ \\
\bottomrule
\end{tabular}
\end{table}

\section{Logistic regression classifier}
\label{app:readout}

In Table~\ref{tab:readout}, we replace the random forest with logistic regression on standardized features for all six configurations. Logistic regression is particularly competitive on TriviaQA, suggesting that classifier choice affects the relative performance of the different configurations.

\begin{table}[t]
\centering
\caption{
Pooled AUROC (\%) for LOTUS and its ablated variants using logistic regression under the question-level split. Results are averaged over five random draws.}
\label{tab:readout}

\resizebox{\columnwidth}{!}{%
\begin{tabular}{lccrrrr}
\toprule
\textbf{Configuration} & \textbf{Summaries} & \textbf{Features}
& \textbf{NQ-Open} & \textbf{TruthfulQA}
& \textbf{TriviaQA} & \textbf{TyDiQA-GP} \\
\midrule

\multicolumn{7}{l}{\textit{Qwen2.5-7B-Instruct}}\\
\textbf{LOTUS (full)}
& $C,S,V$ & $896$
& $69.24$ & $80.16$ & $92.48$ & $\mathbf{84.09}$ \\

\multicolumn{7}{l}{\hspace{0.5em}\textit{Ablated variants}}\\
w/o $S$         & $C,V$ & $576$ & $68.01$ & $81.17$ & $92.56$ & $81.92$ \\
Peak only       & $C$   & $320$ & $\mathbf{71.85}$ & $81.80$ & $92.44$ & $80.99$ \\
Quantile only   & $S$   & $320$ & $70.43$ & $82.07$ & $\mathbf{92.69}$ & $82.82$ \\
Difference only & $V$   & $256$ & $71.45$ & $\mathbf{82.52}$ & $91.69$ & $82.70$ \\
Shared bases    & $S,V$ & $512$ & $68.98$ & $79.39$ & $92.54$ & $82.82$ \\

\midrule

\multicolumn{7}{l}{\textit{LLaMA-3.1-8B}}\\
\textbf{LOTUS (full)}
& $C,S,V$ & $896$
& $79.02$ & $79.38$ & $\mathbf{85.16}$ & $76.97$ \\

\multicolumn{7}{l}{\hspace{0.5em}\textit{Ablated variants}}\\
w/o $S$         & $C,V$ & $576$ & $80.20$ & $80.78$ & $84.71$ & $74.70$ \\
Peak only       & $C$   & $320$ & $\mathbf{81.57}$ & $83.04$ & $84.55$ & $77.09$ \\
Quantile only   & $S$   & $320$ & $81.08$ & $82.48$ & $85.05$ & $\mathbf{78.67}$ \\
Difference only & $V$   & $256$ & $79.56$ & $\mathbf{83.14}$ & $82.22$ & $74.08$ \\
Shared bases    & $S,V$ & $512$ & $78.55$ & $80.93$ & $84.57$ & $73.95$ \\

\bottomrule
\end{tabular}%
}
\end{table}

\section{Rank sensitivity}
\label{app:ranks}

Table~\ref{tab:ranks} reports the rank sensitivity across all thirteen configurations. The peak and range summaries use $r_L=k$, while the update summary uses $r_L=\max(1,k-1)$ to account for its eight-layer structure. It is evident that performance remains relatively stable as the layer rank $k$ varies from $1$ to $9$. The reported setting, $k=5$ and $r_F=64$, remains competitive across different settings.

\begin{table}[h]
\centering\small
\setlength{\tabcolsep}{4pt}
\caption{Rank sensitivity under the question-level split, averaged over five runs. We vary one rank at a time, and width denotes the number of features per response.}
\label{tab:ranks}
\begin{tabular}{lccccc}
\toprule
& & \multicolumn{2}{c}{\textbf{Qwen2.5-7B-Instruct}} & \multicolumn{2}{c}{\textbf{LLaMA-3.1-8B}} \\
\cmidrule(lr){3-4}\cmidrule(lr){5-6}
\textbf{Ranks} & \textbf{Width} & TruthfulQA & TyDiQA-GP & TruthfulQA & TyDiQA-GP \\
\midrule
$k=1$, $r_F=64$ & 192 & 88.04 & 87.41 & 89.07 & 81.32 \\
$k=2$, $r_F=64$ & 320 & 87.88 & 87.36 & 89.82 & 82.75 \\
$k=3$, $r_F=64$ & 512 & 87.63 & 87.78 & 89.45 & 82.81 \\
$k=4$, $r_F=64$ & 704 & 87.89 & 88.03 & 89.33 & 82.59 \\
\textbf{$k=5$, $r_F=64$} & \textbf{896} & \textbf{87.76} & \textbf{87.82} & \textbf{89.22} & \textbf{82.47} \\
$k=6$, $r_F=64$ & 1088 & 87.57 & 87.62 & 89.09 & 82.07 \\
$k=7$, $r_F=64$ & 1280 & 87.78 & 87.55 & 89.12 & 81.74 \\
$k=8$, $r_F=64$ & 1472 & 87.59 & 87.67 & 88.95 & 81.77 \\
$k=9$, $r_F=64$ & 1664 & 87.61 & 87.52 & 88.95 & 81.87 \\
\midrule
$k=5$, $r_F=8$ & 112 & 86.12 & 86.29 & 88.13 & 81.06 \\
$k=5$, $r_F=16$ & 224 & 86.77 & 86.50 & 88.93 & 81.36 \\
$k=5$, $r_F=32$ & 448 & 87.33 & 87.17 & 89.03 & 82.44 \\
$k=5$, $r_F=128$ & 1792 & 87.44 & 87.36 & 89.26 & 82.70 \\
\bottomrule
\end{tabular}
\end{table}

\section{Transfer across models}
\label{app:crossmodel}

To enable cross-model transfer, we align the target model's LOTUS representation with the
feature space expected by the source detector. Let $\bm\phi^{(t)}$ and $\bm\phi^{(s)}$ denote the
LOTUS representations produced by the target and source models, respectively. For each response in
the target model's training set, we first compute $\bm\phi^{(t)}$ using the target model's own
hidden states, scaling parameters, and Tucker bases. We then present the \emph{same response text},
together with its corresponding prompt, to the source model and compute its LOTUS representation
$\bm\phi^{(s)}$ using the source model's bases. This gives an unlabeled paired set
$\{(\bm\phi^{(t)}_j,\bm\phi^{(s)}_j)\}_j$, where the two vectors describe the same
prompt--response pair in the coordinate systems of the two models.

Using these pairs, we fit a ridge regression
$\bm\phi^{(s)} \approx \bm W\bm\phi^{(t)}+\bm b$
that maps target-model LOTUS features into the source model's feature space. Importantly, this
alignment uses only paired representations and does not use whether a response is hallucinated or
correct. At test time, a target-model response is represented only in the target LOTUS space,
mapped through the learned ridge transformation, and then passed directly to the source model's
pretrained hallucination detector. Thus, the source detector remains fixed and no target-model
hallucination labels are required.}

We train the detector separately for each model in Section~\ref{sec:method}, since the scaling statistics and Tucker-2 bases are specific to that model's hidden-state space. Its robust scaling and its Tucker-2
bases live in that model's hidden space, and the bases are defined only up to sign, order and rotation, so the coordinates the
random forest reads have no counterpart in another model. We therefore evaluate whether a detector trained on one source model can score responses from another target model without using any labels.

\paragraph{Protocol.} 
For each question-level split, the source detector is trained as in Section~\ref{sec:readout}, using only questions that appear in the training splits of both models. Target responses are represented using the target model's own scaling and Tucker-2 bases, fitted without target labels.  The target's answers are projected with the target model's own scaling and bases,
which are fitted without labels on the target's training answers, as for the target's own detector. To
carry these coordinates into the source's, the source model reads the target's training answers: each
answer is appended to the source model's own prompt for the same question, and the window layer features of
Section~\ref{sec:summaries} are pooled over the answer's tokens and projected with the source's bases.
A ridge regression from target to source coordinates is fitted on these pairs, with the penalty chosen
by question-level five-fold cross-validation. No target labels are used during alignment or training. Table~\ref{tab:crossmodel_main} compares the target-specific detector, cross-model transfer with and without alignment, and the source detector applied to
the source model's hidden states while it reads the target's answers.

}

\section{Computational Cost}
\label{app:cost}

In Table~\ref{tab:cost},  we compare the storage required per response for the three trained detectors using uncompressed float16 representations across both model sizes. It is evident that LOTUS is substantially more storage-efficient than ACT-ViT across all datasets. At $N=20$, ACT-ViT requires roughly $3$--$4\times$ more storage, and its default $N=100$ increases this gap to about $19\times$. The difference is largest on TriviaQA, where LOTUS uses $28.6$--$32.7$ GB compared with $106$--$122$ GB for ACT-ViT at $N=20$. HARP avoids stored activations entirely by recomputing the final-layer states during evaluation. Thus, LOTUS provides a much smaller memory footprint while retaining a fixed-size representation for each response.

\begin{table}[h]
\centering\small
\setlength{\tabcolsep}{5pt}
\caption{Stored input (GB) per model and dataset, float16 and uncompressed, as a range over
Qwen2.5-7B-Instruct and LLaMA-3.1-8B. $N$ denotes ACT-ViT's number of tokens, where we report $N=20$ and the released default is $N=100$.}
\label{tab:cost}
\begin{tabular}{lcccc}
\toprule
\textbf{Detector} & \textbf{TruthfulQA} & \textbf{TyDiQA-GP} & \textbf{NQ-Open} & \textbf{TriviaQA} \\
\midrule
HARP & \multicolumn{4}{c}{none: the last layer's states are recomputed in its own script} \\
ACT-ViT, $N=20$ & $8.7$--$10.0$ & $4.7$--$5.4$ & $38.6$--$44.1$ & $106$--$122$ \\
ACT-ViT, $N=100$ & $43.6$--$49.9$ & $23.5$--$26.9$ & $193$--$220$ & $532$--$608$ \\
\textbf{LOTUS} & $2.3$--$2.7$ & $1.3$--$1.4$ & $10.4$--$11.8$ & $28.6$--$32.7$ \\
\bottomrule
\end{tabular}
\end{table}

\section{Token Budget Analysis of ACT-ViT}
\label{app:neff}

ACT-ViT max-pools its activation tensor to a fixed number of tokens $N$, with $N=100$ as the released default. Since our generated responses are short, we use $N=20$ in the main experiments to reduce unnecessary padding and storage. 
In Table ~\ref{tab:neff}, we compare the performance of $N=20$ and $N=100$ for Qwen2.5-7B-Instruct using both pooled and within-question AUROC. The results show that increasing $N$=$100$ yields only modest improvements on TruthfulQA and TyDiQA-GP, while increasing activation storage by approximately $5\times$ more. LOTUS remains ahead of ACT-ViT under the question-level split at both token settings.%

\begin{table}[h]
\centering\small
\setlength{\tabcolsep}{4pt}
 \caption{ACT-ViT performance across different numbers of tokens $N$. We report pooled AUROC (\%) under question-level (Q) and answer-level (A) splits, and within-question AUROC under the question-level split, averaged over five runs with standard deviations.}
\label{tab:neff}
\begin{tabular}{lcccc}
\toprule
\textbf{Model, dataset} & $N$ & \textbf{Pooled, Q} & \textbf{Pooled, A} & \textbf{Within, Q} \\
\midrule
Qwen2.5-7B-Instruct, TruthfulQA & $20$ & $85.40 \pm 1.01$ & $87.58 \pm 0.56$ & $78.51 \pm 4.51$ \\
Qwen2.5-7B-Instruct, TruthfulQA & $100$ & $85.65 \pm 1.74$ & $88.18 \pm 0.59$ & $79.26 \pm 3.89$ \\
Qwen2.5-7B-Instruct, TyDiQA-GP & $20$ & $82.82 \pm 1.46$ & $86.81 \pm 1.69$ & $81.93 \pm 1.93$ \\
Qwen2.5-7B-Instruct, TyDiQA-GP & $100$ & $84.05 \pm 1.52$ & $89.93 \pm 2.12$ & $85.32 \pm 2.43$ \\
LLaMA-3.1-8B, TruthfulQA & $20$ & $89.07 \pm 1.30$ & $90.68 \pm 0.60$ & $83.50 \pm 3.12$ \\
LLaMA-3.1-8B, TyDiQA-GP & $20$ & $79.43 \pm 1.41$ & $84.72 \pm 1.47$ & $80.50 \pm 2.34$ \\
\bottomrule
\end{tabular}
\end{table}

\section{Performance Across Splitting Protocols}
\label{app:std}

We report the performance of HARP, ACT-ViT, and LOTUS under both question-level and answer-level splits, including the standard deviation over five runs in Table~\ref{tab:std-main}. Across most settings, the variability is relatively small, while the answer-level split generally shows lower standard deviations than the question-level split.  However, for Qwen2.5-7B-Instruct on NQ-Open dataset under the question-level split, the standard deviation increases to $5.24\%$ because the test set contains only about $280$ correct responses. 

\begin{table}[h]
\centering\small
\setlength{\tabcolsep}{4pt}
\caption{HARP, ACT-ViT and LOTUS under the question-level (Q) and answer-level (A) splits: pooled AUROC (\%), mean $\pm$ standard deviation over five runs.}
\label{tab:std-main}
\begin{tabular}{lcccc}
\toprule
\textbf{Method, split} & \textbf{NQ-Open} & \textbf{TruthfulQA} & \textbf{TriviaQA} & \textbf{TyDiQA-GP} \\
\midrule
\multicolumn{5}{l}{\textit{Qwen2.5-7B-Instruct}} \\
HARP, Q & $67.73 \pm 2.24$ & $77.54 \pm 1.51$ & $88.75 \pm 0.33$ & $79.90 \pm 2.87$ \\
HARP, A & $81.03 \pm 2.41$ & $85.16 \pm 1.23$ & $92.50 \pm 0.24$ & $91.75 \pm 0.95$ \\
LOTUS, Q & $79.29 \pm 5.24$ & $87.83 \pm 0.89$ & $91.69 \pm 0.10$ & $87.86 \pm 1.95$ \\
LOTUS, A & $91.14 \pm 1.01$ & $94.03 \pm 0.44$ & $94.05 \pm 0.10$ & $94.72 \pm 0.74$ \\
ACT-ViT, Q & $79.11 \pm 5.80$ & $85.40 \pm 1.01$ & $92.77 \pm 0.28$ & $82.82 \pm 1.46$ \\
ACT-ViT, A & $82.92 \pm 2.82$ & $87.58 \pm 0.56$ & $93.59 \pm 0.14$ & $86.81 \pm 1.69$ \\
\midrule
\multicolumn{5}{l}{\textit{LLaMA-3.1-8B}} \\
HARP, Q & $80.56 \pm 1.06$ & $79.58 \pm 5.38$ & $82.54 \pm 0.49$ & $74.99 \pm 0.88$ \\
HARP, A & $86.56 \pm 0.44$ & $84.84 \pm 1.91$ & $88.08 \pm 0.29$ & $85.65 \pm 0.78$ \\
LOTUS, Q & $85.36 \pm 0.42$ & $89.19 \pm 1.41$ & $84.69 \pm 0.20$ & $82.49 \pm 1.23$ \\
LOTUS, A & $91.96 \pm 0.24$ & $94.41 \pm 0.58$ & $89.48 \pm 0.10$ & $90.07 \pm 0.58$ \\
ACT-ViT, Q & $82.98 \pm 1.76$ & $89.07 \pm 1.30$ & $87.80 \pm 0.33$ & $79.43 \pm 1.41$ \\
ACT-ViT, A & $87.39 \pm 0.61$ & $90.68 \pm 0.60$ & $89.66 \pm 0.23$ & $84.72 \pm 1.47$ \\
\bottomrule
\end{tabular}
\end{table}

\begin{table}[h]
\centering\small
\setlength{\tabcolsep}{4pt}
\caption{HARP, both ACT-ViT variants and LOTUS on TruthfulQA and TyDiQA-GP under the question-level (Q) and answer-level (A) splits, with standard deviations: pooled AUROC (\%), mean $\pm$ standard deviation over five draws.}
\label{tab:std-actvit}
\begin{tabular}{lcccc}
\toprule
& \multicolumn{2}{c}{\textbf{Qwen2.5-7B-Instruct}} & \multicolumn{2}{c}{\textbf{LLaMA-3.1-8B}} \\
\cmidrule(lr){2-3}\cmidrule(lr){4-5}
\textbf{Method, split} & TruthfulQA & TyDiQA-GP & TruthfulQA & TyDiQA-GP \\
\midrule
HARP, Q & $77.54 \pm 1.51$ & $79.90 \pm 2.87$ & $79.58 \pm 5.38$ & $74.99 \pm 0.88$ \\
HARP, A & $85.16 \pm 1.23$ & $91.75 \pm 0.95$ & $84.84 \pm 1.91$ & $85.65 \pm 0.78$ \\
ACT-ViT (joint), Q & $83.13 \pm 1.18$ & $77.70 \pm 2.99$ & $86.78 \pm 1.20$ & $77.21 \pm 1.84$ \\
ACT-ViT (joint), A & $87.79 \pm 0.36$ & $83.88 \pm 2.29$ & $90.14 \pm 0.56$ & $81.38 \pm 2.80$ \\
ACT-ViT, Q & $85.40 \pm 1.01$ & $82.82 \pm 1.46$ & $89.07 \pm 1.30$ & $79.43 \pm 1.41$ \\
ACT-ViT, A & $87.58 \pm 0.56$ & $86.81 \pm 1.69$ & $90.68 \pm 0.60$ & $84.72 \pm 1.47$ \\
LOTUS, Q & $87.83 \pm 0.89$ & $87.86 \pm 1.95$ & $89.19 \pm 1.41$ & $82.49 \pm 1.23$ \\
LOTUS, A & $94.03 \pm 0.44$ & $94.72 \pm 0.74$ & $94.41 \pm 0.58$ & $90.07 \pm 0.58$ \\
\bottomrule
\end{tabular}
\end{table}

\section{Within-question AUROC}
\label{app:within}

{\bf Definition.}
For each question $i$, let
\[
\mathcal H_i=\{k:z_{i,k}=1\},
\qquad
\mathcal C_i=\{k:z_{i,k}=0\}
\]
denote the sets of hallucinated and correct sampled responses,
respectively. We consider only questions containing at least one
response of each class,
\[
\mathcal Q_{\mathrm{mix}}
=
\{i:|\mathcal H_i|>0,\;|\mathcal C_i|>0\}.
\]
Given a hallucination score $s_{i,k}$, where larger values indicate
stronger evidence of hallucination, we define the within-question
AUROC as
\begin{equation}
\mathrm{AUROC}_{\mathrm{WQ}}
=
\frac{1}{N_{\mathrm{pair}}}
\sum_{i\in\mathcal Q_{\mathrm{mix}}}
\sum_{k\in\mathcal H_i}
\sum_{k'\in\mathcal C_i}
\left[
\mathbb{I}(s_{i,k}>s_{i,k'})
+
\frac{1}{2}\mathbb{I}(s_{i,k}=s_{i,k'})
\right],
\label{eq:wq-auroc}
\end{equation}
where
\[
N_{\mathrm{pair}}
=
\sum_{i\in\mathcal Q_{\mathrm{mix}}}
|\mathcal H_i|\,|\mathcal C_i|.
\]

}

We report within-question AUROC under the question-level split for LOTUS, HARP, and ACT-ViT in Table~\ref{tab:within}. The remaining baselines assign one score per question, so their within-question AUROC is always $50\%$ and does not distinguish their responses. LOTUS outperforms HARP in all four settings and ACT-ViT in three of the four. On TruthfulQA, HARP shows a noticeably larger drop from pooled to within-question AUROC, suggesting that more of its pooled performance is driven by differences between questions. LOTUS and ACT-ViT show smaller gaps, indicating better separation among responses to the same question. On TyDiQA-GP, pooled and within-question AUROC performance are much closer for all three detectors. %

\begin{table}[h]
\centering\small
\setlength{\tabcolsep}{4pt}
\caption{Within-question AUROC (\%) of the three detectors that score each answer, under the question-level split: mean $\pm$ standard deviation over five draws. LOTUS and ACT-ViT share the same test sets; HARP uses the draws of its own code. ACT-ViT uses $N=20$.}
\label{tab:within}
\begin{tabular}{lcccc}
\toprule
& \multicolumn{2}{c}{\textbf{Qwen2.5-7B-Instruct}} & \multicolumn{2}{c}{\textbf{LLaMA-3.1-8B}} \\
\cmidrule(lr){2-3}\cmidrule(lr){4-5}
\textbf{Method} & TruthfulQA & TyDiQA-GP & TruthfulQA & TyDiQA-GP \\
\midrule
\textbf{LOTUS} & $82.07 \pm 2.18$ & $88.09 \pm 3.77$ & $82.73 \pm 1.92$ & $81.34 \pm 3.01$ \\
HARP & $67.84 \pm 2.07$ & $82.69 \pm 1.44$ & $67.78 \pm 8.05$ & $75.87 \pm 1.63$ \\
ACT-ViT & $78.51 \pm 4.51$ & $81.93 \pm 1.93$ & $83.50 \pm 3.12$ & $80.50 \pm 2.34$ \\
\midrule
Answer pairs per draw & 1{,}820--2{,}059 & 940--1{,}038 & 1{,}840--1{,}942 & 1{,}576--1{,}768 \\
\quad HARP's draws & 1{,}896--2{,}137 & 924--1{,}086 & 1{,}728--1{,}930 & 1{,}524--1{,}719 \\
\bottomrule
\end{tabular}
\end{table}

\section{Reproduction of HARP}
\label{app:harp}

We ran the released implementation of HARP \citep{hu2025harp} on our generations and labels
(Section ~\ref{sec:setup}). Our answers replace its cache of generations, so its own generation and labeling
code does not run, and the only modification is to two device strings. Table~\ref{tab:harp} compares
the published results, obtained on the authors' generations with the answer-level split of the released
code, with our runs. For Qwen2.5-7B-Instruct, a single run of the released script differs from the
published results by $-1.83$ to $+3.47$ points, $+0.21$ on average. For LLaMA-3.1-8B, five draws of the
released split fall below the published results on every dataset, by $0.95$ to $4.82$ points and $3.07$
on average; we have not identified the cause. Under the question-level split used throughout the paper,
the same detector is $3.75$ to $13.30$ points lower than under the released split
(Section ~\ref{sec:res-leak}).

\begin{table}[h]
\centering\small
\setlength{\tabcolsep}{4pt}
\caption{HARP with $256$ directions: published AUROC (\%) on the authors' generations, and our runs of the released code on our generations and labels. The released split is the answer-level split; five draws are reported as mean $\pm$ standard deviation.}
\label{tab:harp}
\begin{tabular}{lcccc}
\toprule
\textbf{Result} & \textbf{NQ-Open} & \textbf{TruthfulQA} & \textbf{TriviaQA} & \textbf{TyDiQA-GP} \\
\midrule
\multicolumn{5}{l}{\textit{Qwen2.5-7B-Instruct}} \\
Published & 84.0 & 88.1 & 92.8 & 88.4 \\
Released split, 1 run & 83.12 & 86.27 & 92.89 & 91.87 \\
Released split, 5 draws & $81.03 \pm 2.41$ & $85.16 \pm 1.23$ & $92.50 \pm 0.24$ & $91.75 \pm 0.95$ \\
Question-level split, 5 draws & $67.73 \pm 2.24$ & $77.54 \pm 1.51$ & $88.75 \pm 0.33$ & $79.90 \pm 2.87$ \\
\midrule
\multicolumn{5}{l}{\textit{LLaMA-3.1-8B}} \\
Published & 89.4 & 88.5 & 92.9 & 86.6 \\
Released split, 5 draws & $86.56 \pm 0.44$ & $84.84 \pm 1.91$ & $88.08 \pm 0.29$ & $85.65 \pm 0.78$ \\
Question-level split, 5 draws & $80.56 \pm 1.06$ & $79.58 \pm 5.38$ & $82.54 \pm 0.49$ & $74.99 \pm 0.88$ \\
\bottomrule
\end{tabular}
\end{table}

\section{Agreement with the labels of HARP}
\label{app:labels}

Our labels follow the rubric of \citet{hu2025harp}: an answer is correct if its ROUGE-L value with the reference
answers is at least $0.7$ or its BLEURT-20 score against a reference exceeds $0.5$. We average the ROUGE-L scores across all reference answers and compare the average with $0.7$, while the released judge uses the highest ROUGE-L score and compares it with $0.7$. We use the same BLEURT-20 criterion in both implementations, where the highest score across the reference answers is compared with a threshold value of $0.5$. Table~\ref{tab:labels} compares the labels from the released judge with ours on all $148{,}270$ Qwen2.5-7B-Instruct responses. The two labeling procedures agree on $99.27\%$--$99.94\%$ of responses across datasets, and only $44$ of the $14{,}827$ questions change between known and unknown status. %

\begin{table}[h]
\centering\small
\setlength{\tabcolsep}{4pt}
\caption{Our labels against those of the released judge of HARP, on the answers of Qwen2.5-7B-Instruct. \emph{Changed} counts questions whose known or unknown status differs between the two sets of labels; the last two columns give the agreement (\%) over answers and over the known or unknown status of questions.}
\label{tab:labels}
\begin{tabular}{lrrrrrr}
\toprule
& & \multicolumn{2}{c}{\textbf{Known questions}} & & \multicolumn{2}{c}{\textbf{Agreement}} \\
\cmidrule(lr){3-4}\cmidrule(lr){6-7}
\textbf{Dataset} & \textbf{Questions} & Ours & HARP & \textbf{Changed} & Answers & Status \\
\midrule
NQ-Open & $3{,}610$ & $360$ & $358$ & $4$ & $99.94$ & $99.89$ \\
TruthfulQA & $817$ & $666$ & $670$ & $6$ & $99.57$ & $99.27$ \\
TriviaQA & $9{,}960$ & $6{,}316$ & $6{,}331$ & $33$ & $99.71$ & $99.67$ \\
TyDiQA-GP & $440$ & $302$ & $301$ & $1$ & $99.27$ & $99.77$ \\
\midrule
Total & $14{,}827$ & $7{,}644$ & $7{,}660$ & $44$ & & \\
\bottomrule
\end{tabular}
\end{table}

\section{Covariance spectra}
\label{app:spectrum}

Table~\ref{tab:spectrum} reports the values summarized in Section~\ref{sec:results} for the embedding output, our selected layer window, and the final block, together with the variance captured by the leading $64$ directions. Hallucinated responses consistently have higher effective rank than correct responses, while both classes retain a large share of their variance in a low-dimensional subspace. The selected window also achieves substantially higher single-layer AUROC than the embedding and final block.

\begin{table}[h]
\centering\small
\setlength{\tabcolsep}{4pt}
\caption{Covariance spectra of peak summaries for correct (C) and hallucinated (H) responses from Qwen2.5-7B-Instruct on TyDiQA-GP, averaged over $20$ runs. We report size-matched and question-paired comparisons for the embedding, blocks $15$--$23$, and the final block. The last column reports single-layer AUROC (\%) using logistic regression with question-grouped five-fold cross-validation (Appendix~\ref{app:window}).}
\label{tab:spectrum}
\begin{tabular}{llccccc}
\toprule
& & \multicolumn{2}{c}{\textbf{Effective rank}} & \multicolumn{2}{c}{\textbf{Top-64 variance}} & \\
\cmidrule(lr){3-4}\cmidrule(lr){5-6}
\textbf{Draw} & \textbf{Layer} & C & H & C & H & \textbf{AUROC} \\
\midrule
Size-matched & Embedding             & $47.3$  & $136.6$ & $0.715$ & $0.575$ & $76.5$ \\
             & Blocks $15$--$23$     & $97.7$  & $204.7$ & $0.653$ & $0.538$ & $86.8$ \\
             & Block $27$ (last)     & $109.7$ & $215.1$ & $0.626$ & $0.520$ & $83.3$ \\
\midrule
Paired       & Embedding             & $35.0$  & $66.5$  & $0.780$ & $0.684$ & $76.5$ \\
             & Blocks $15$--$23$     & $59.6$  & $93.4$  & $0.716$ & $0.650$ & $86.8$ \\
             & Block $27$ (last)     & $60.2$  & $95.4$  & $0.700$ & $0.628$ & $83.3$ \\
\bottomrule
\end{tabular}
\end{table}

\end{document}